\documentclass[11pt, a4paper, twocolumn, copyright, google]{google}

\usepackage[authoryear, sort&compress, round]{natbib}
\uselogo{} 

\usepackage{tcolorbox}

\definecolor{boxblue}{HTML}{c6dbfc}
\definecolor{boxyellow}{HTML}{ffe5b2}
\definecolor{boxred}{HTML}{f3c0bd}

\usepackage{xcolor} 

\usepackage[dvipsnames]{xcolor}

\usepackage{subcaption}
\usepackage{algorithm}
\usepackage{algpseudocode}
\usepackage{amsmath}
\usepackage{amssymb}
\usepackage{nicefrac}
\usepackage{comment}
\usepackage{booktabs}
\usepackage{xcolor}
\usepackage{colortbl}
\usepackage{hyperref}
\usepackage{etoc}
\usepackage{multirow}
\usepackage{enumitem}
\usepackage{adjustbox}
\usepackage{amsmath}
\usepackage{tikz}
\usetikzlibrary{arrows.meta, decorations.pathreplacing, calc}
\usepackage{nicefrac}
\usepackage{array}
\usepackage{tabularx}
\usepackage{array}
\newcolumntype{C}{>{\centering\arraybackslash}p{1.1cm}} 

\renewcommand*{\backref}[1]{}
\renewcommand*{\backrefalt}[4]{%
    \ifcase #1 (Not cited.)%
    \or        (Cited on page #2.)%
    \else      (Cited on pages #2.)%
    \fi}

\usepackage{placeins} 

\usepackage{cuted}

\usepackage{fancyhdr}
\usepackage{xspace}

\algblock{ParFor}{EndParFor}
\algnewcommand\algorithmicparfor{\textbf{parallel for}}
\algnewcommand\algorithmicpardo{\textbf{do}}
\algnewcommand\algorithmicendparfor{\textbf{end\ parallel for}}
\algrenewtext{ParFor}[1]{\algorithmicparfor\ #1\ \algorithmicpardo}
\algrenewtext{EndParFor}{\algorithmicendparfor}

\newcommand{\mergefn}{\texttt{Merge}\xspace}

\DeclareMathOperator{\rda}{RDA}

\title{\LARGE Decoupled DiLoCo for Resilient Distributed Pre-training
}

\renewcommand{\today}{}

\author[1,2]{Decoupled DiLoCo Team}

\affil[1]{Google DeepMind}
\affil[2]{Google Research}

\begin{abstract}
Modern large-scale language model pre-training relies heavily on the single program multiple data (SPMD) paradigm, which requires tight coupling across accelerators. Due to this coupling, transient slowdowns, hardware failures, and synchronization overhead stall the entire computation, wasting significant compute time at scale. 
While recent distributed methods like DiLoCo reduced communication bandwidth, they remained fundamentally synchronous and vulnerable to these system stalls.
To address this, we introduce Decoupled DiLoCo, an evolution of the DiLoCo framework designed to break the lock-step synchronization barrier and go beyond SPMD to maximize training goodput.
Decoupled DiLoCo partitions compute across multiple independent ``learners'' that execute local inner optimization steps.
These learners asynchronously communicate parameter fragments to a central synchronizer, which circumvents failed or straggling learners by aggregating updates using a minimum quorum, an adaptive grace window, and dynamic token-weighted merging.
Inspired by ``chaos engineering'', we achieve significantly improved training efficiency in failure-prone environments with millions of simulated chips with strictly zero global downtime, while maintaining competitive model performance across text and vision tasks, for both dense and mixture-of-expert architectures.
\end{abstract}

\begin{document}

\maketitle

\section{Introduction}\label{sec:intro}

Modern LLM pre-training relies on the tightly coupled single program, multiple data (SPMD) paradigm (e.g., data, tensor, and sequence parallelism) that requires global synchronization at every step. This ``monolithic'' approach creates a major reliability bottleneck: a single hardware failure or straggler can stall the entire system. As compute scales, the sheer number of components transforms rare hardware failures into routine occurrences. This is exacerbated by the long duration of pre-training regimes, where frequent interruptions lead to significant downtime and wasted compute.

\begin{figure*}[!t]
    \centering
    \captionsetup{justification=centering}
    \includegraphics[width=\textwidth, trim={0cm 1cm 0cm 0cm}, clip]{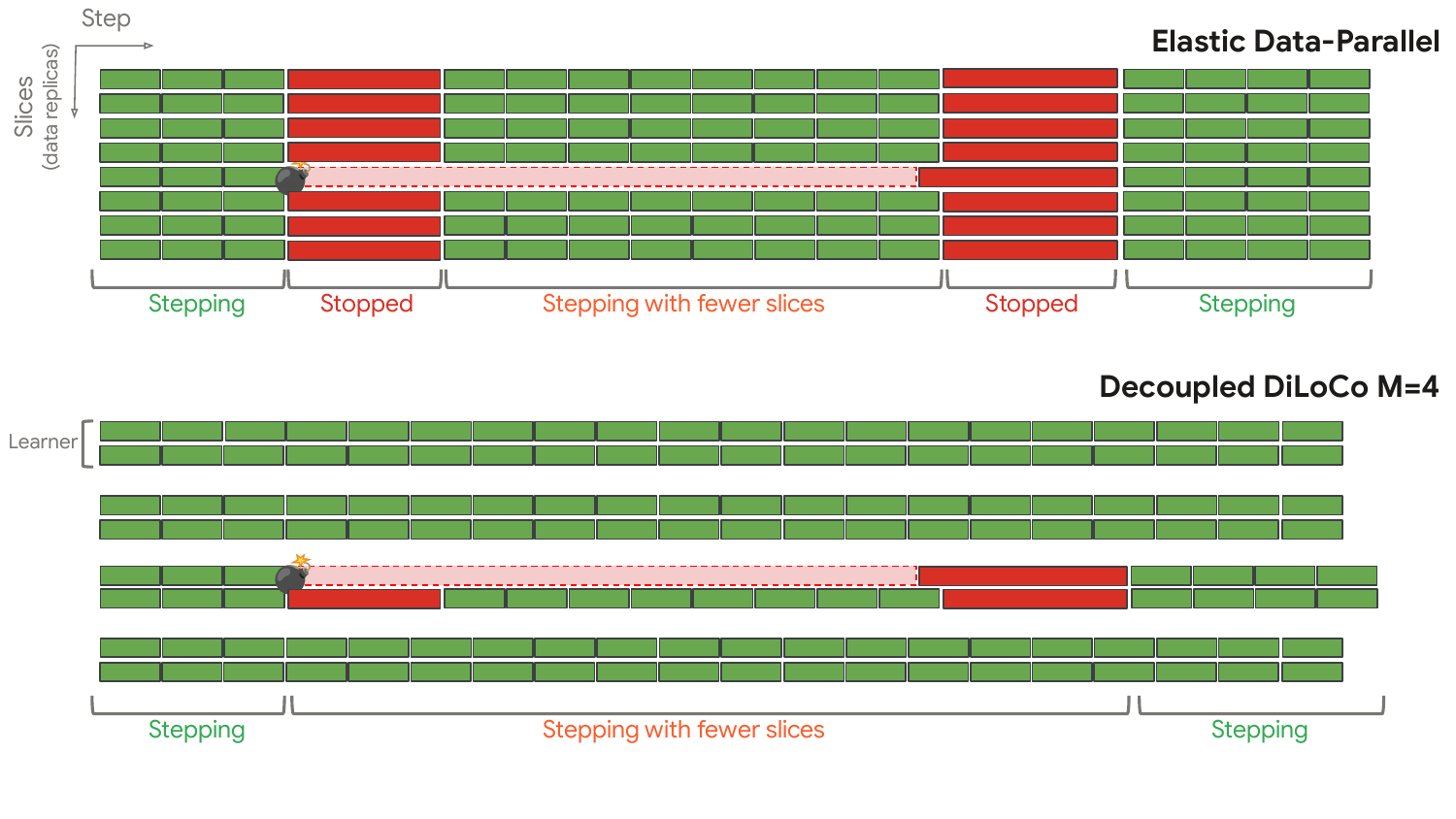}
    \caption{\textbf{Slice-granularity elasticity vs decoupling}: the system continues training with fewer ``slices'' of TPU chips when there is a localized failure. When a failure occurs affecting 1 of the $M$ replicas, the decoupling approach allows the other $\nicefrac{M-1}{M}$ replicas in the system to continue stepping.
    }
    \label{fig:combined_elastic}
\end{figure*}

Framing this analogously to the CAP theorem~\citep{brewer2000towards}, we argue the primary bottleneck of modern pre-training is a rigid adherence to parameter consistency. To reason about this trade-off, we define an analog of the ``CAP properties'' for the pre-training setting as follows:
\begin{itemize}[leftmargin=5pt, nosep]
    \item \textbf{Consistency (C):} Every accelerator maintains a view of a globally synchronized set of model weights.
    \item \textbf{Availability (A):} Training continues in the presence of hardware failures.
    \item \textbf{Partition Tolerance (P):} Training continues despite interconnect instability or communication delays.
\end{itemize}
Under this view, SPMD pre-training prioritizes consistency over all else: it sacrifices availability and partition tolerance to ensure every accelerator maintains a view of the global model state. While recent slice-granular ``elastic'' methods \citep{geminiteam2025gemini2p5} can reconfigure the SPMD computation to run on a smaller subset of healthy accelerators, the global overhead of fault detection and cluster-wide resizing still incurs significant downtime, as illustrated in \autoref{fig:combined_elastic} (top).

Recent distributed alternatives like DiLoCo~\citep{douillard2024diloco} and its streaming variant~\citep{douillard2025streaming} have successfully addressed communication bottlenecks by reducing bandwidth requirements through intermittent synchronization. However, because previous iterations of DiLoCo remained fundamentally synchronous, they still enforced strict consistency across the cluster, leaving the system just as vulnerable to localized hardware failures and straggler penalties.

Inspired by the Pathways vision~\citep{barham2022pathways}, we argue that pre-training should prioritize availability and partition tolerance over consistency by moving beyond the present tightly-coupled paradigm. We propose Decoupled DiLoCo, a distributed training framework that decomposes a global cluster into independent, asynchronous ``learners''. By evolving DiLoCo's intermittent synchronization into a fully asynchronous communication protocol, we limit the ``blast radius'' of a hardware failure to a single learner. This allows the majority of the system to continue training uninterrupted during local failures or reconfigurations (\autoref{fig:combined_elastic}). Through careful system and algorithm co-design, we demonstrate that maximum training goodput can be achieved in chaotic environments without sacrificing final model quality.

\paragraph{Contributions}

Our core contributions are:
\begin{itemize}
    \item \textbf{Decoupled DiLoCo:} We introduce a distributed training framework that evolves previous bandwidth-focused methods by decomposing monolithic SPMD clusters into independent, asynchronous learners. This prioritizes availability and partition tolerance over strict consistency.
    \item \textbf{Scalable System Architecture:} We design a system architecture featuring a central synchronizer to facilitate asynchronous parameter reconciliation, enabling the framework to operate efficiently at pre-training scales.
    \item \textbf{Empirical Validation:} We provide extensive empirical evidence that Decoupled DiLoCo achieves comparable downstream performance to standard data-parallel training across various model and compute scales for both dense and Mixture-of-Experts (MoE) architectures on text and multi-modal evaluations. Moreover we show that pre-training with our asynchronous framework doesn't hinder later post-training capabilities.
    \item \textbf{Robustness via Chaos Engineering:} We apply chaos engineering~\citep{basiri2016chaos} principles to LLM pre-training, demonstrating that our framework maintains high availability and model quality, even under aggressive, continuous hardware failures.
\end{itemize}

\section{Preliminaries}\label{sec:preliminaries}

Let $\theta^{(t)}$ denote the model parameters at step $t$. We wish to train a model across $M$ compute clusters (called \emph{learners} for brevity), each of which has its own model replica $\theta_m^{(t)}$, for $m \in [M]$. In SPMD data-parallel training, each learner computes a batched gradient, and a global gradient is aggregated across learners. This requires synchronization across the learners at every step, incurring potentially high amounts of communication and bandwidth. It also means that a slowdown in a single learner delays the entire computation, and a failure requires, in principle, a replay of the partially-completed step.

In DiLoCo~\citep{douillard2024diloco}, each of the $M$ learners trains in parallel using an \emph{inner optimizer} (e.g. AdamW) on distinct data shards, and only synchronize weights every $H$ steps. Rather than averaging the weights, DiLoCo applies an \emph{outer optimization} step~\citep{reddi2021adaptive} when $t \bmod H = 0$ by treating differences in parameter space as gradient estimates:
\begin{equation}\label{eq:outer_opt}
\begin{aligned}
    \Delta^{(t)} &= \dfrac{1}{M}\sum_{m=1}^M \Delta_m^{(t)} = \dfrac{1}{M}\sum_{m=1}^M\left(\theta_m^{(t - H)} - \theta_m^{(t)}\right) \\
    \theta^{(t)}_m &= \texttt{OuterOpt}(\theta^{(t-H)}_m, \Delta^{(t)})
\end{aligned}
\end{equation}

We refer to $\Delta_m^{(t)}$ as an \emph{outer gradient}. \citet{douillard2024diloco} found that letting \texttt{OuterOpt} be SGD with Nesterov momentum~\citep{sutskever2013nesterov} greatly improves model quality compared to weight averaging (which corresponds to an outer optimizer of SGD with learning rate 1).

While DiLoCo reduces total bandwidth usage by a factor of $H$, it does not reduce peak bandwidth. This can be achieved by Streaming DiLoCo~\citep{douillard2025streaming}. Partition the model weights $\theta$ into $P$ sets (called \emph{fragments}), and let $\theta_{m, p}^{(t)}$ denote the $p$-th fragment held by learner $m$ at step $t$. Let $t_1, t_2, \dots, t_P \in \{0, \dots, H-1\}$ be distinct offsets. Then, at step $t$ we only synchronize a fragment $p$ if $t\bmod H = t_p$.

Since \eqref{eq:outer_opt} requires no actual gradient, the outer gradients can be computed fragment-wise. For $t \bmod H = t_p$, we compute
\begin{equation}\label{eq:streaming_outer_opt}
\begin{aligned}
    \Delta^{(t)}_p &= \dfrac{1}{M}\sum_{m=1}^M \Delta_{m, p}^{(t)} = \dfrac{1}{M}\sum_{m=1}^M\left(\theta_{m, p}^{(t - H)} - \theta_{m, p}^{(t)}\right) \\
    \theta^{(t)}_{m, p} &= \texttt{OuterOpt}(\theta^{(t-H)}_{m, p}, \Delta^{(t)}_p)
\end{aligned}
\end{equation}
As written, \eqref{eq:streaming_outer_opt} is blocking - the learners can't continue training until they send $\Delta_{m, p}^{(t)}$ and receive $\Delta_p^{(t)}$. However, \citet{douillard2025streaming} show that this communication and training can be overlapped - the learners can send $\Delta_{m, p}^{(t)}$ and receive $\Delta_p^{(t)}$ $\tau$ steps later with little change in model quality for small values of $\tau$.

Streaming DiLoCo offers a number of benefits, including significant reductions in bandwidth usage (total and peak) and the ability to make communication across steps asynchronous. However, Streaming DiLoCo still requires lock-step training across learners. Thus, when using Streaming DiLoCo with all-reduce, issues such as stragglers and learner failures can significantly slow down or halt training entirely.

\section{Decoupled DiLoCo}\label{sec:decoupled_diloco}

To break this lock-step barrier, we completely decouple the learners from one another. We now describe Decoupled DiLoCo in full. We train across $M$ learners, each with their own model copy $\theta_m$. As in Streaming DiLoCo, we partition the model across non-overlapping fragments $\{\theta_{m, p}\}_{p \in \mathcal{P}}$, where $P = |\mathcal{P}|$. We employ the same fragment-wise outer optimization as in \eqref{eq:streaming_outer_opt}. However, instead of a blocking global exchange, this optimization is performed by a central synchronizer (the \emph{syncer}), which asynchronously receives and sends updates to learners. We discuss the learners and syncer in detail below.

\begin{figure*}[!t]
    \centering
    \captionsetup{justification=centering}
    \includegraphics[width=\textwidth, trim={0cm 7.5cm 7.5cm 0cm}, clip]{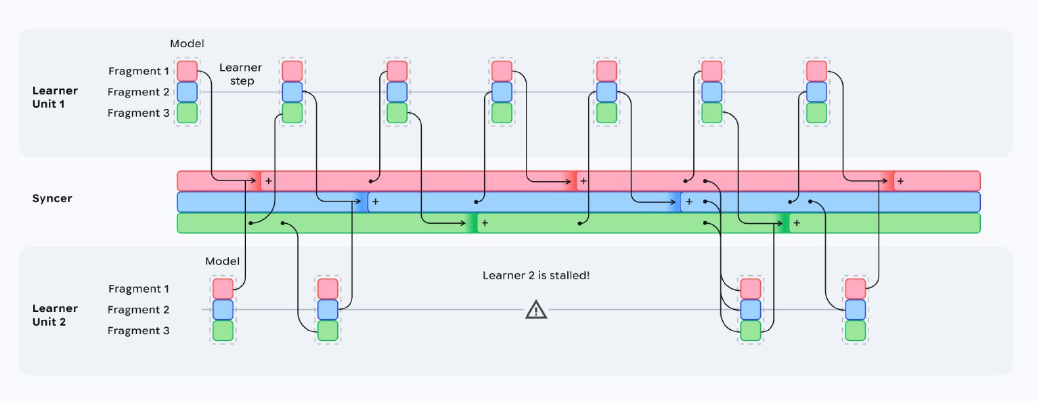}
    \caption{\textbf{Decoupled DiLoCo}. For illustrative purposes, a simple example with $M=2$ learners, $P=3$ fragments, synchronized at every step ($H=3$), and overlapped over $\tau=2$ steps. The second learner stalls for three steps, but the overall training never stops. All missed updates are applied to the faulty learner's state once it continues training.}
    \label{fig:main_decoupled}
\end{figure*}

\subsection{The Learner: Local Optimization}\label{sec:decoupled_learner}

In \autoref{alg:learner}, each learner $m$ operates independently on its own data shard $\mathcal{D}_m$ (\texttt{L8}), without waiting for its peers. A learner continuously executes inner optimization steps (\texttt{L11)} using its inner optimizer (e.g., AdamW). Because learners are decoupled, they operate at varying speeds and may encounter transient failures. To account for this, we track the local progress by maintaining local counters for the number of steps $t_m$ taken by the learner, the number of steps $c^{\text{steps}}_{m, p}$ taken since it received an update to fragment $p$, and the number of tokens $c^{\text{tokens}}_{m, p}$ processed since it received an update to fragment $p$.

At each step, the learner sends its metadata $(t_m, \{c^{\text{steps}}_{m, p}\}_{p \in \mathcal{P}}, \{c^{\text{tokens}}_{m, p}\}_{p\in \mathcal{P}})$ to the syncer (\texttt{L13}).
The syncer uses this to drive the merging of fragments across learners, pulling fragments as needed.
Crucially, this communication occurs in the background, while the learner continues to perform optimization steps. By never waiting for the syncer or its peers, the learner maintains high local goodput and completely isolates the blast radius of any localized hardware failure. Concurrently, the learner listens for updates from the syncer. Upon receiving an updated fragment $\Theta_p$ (\texttt{L15}), the learner overwrites its local weights for that fragment and resets $c^{\text{steps}}_{m, p}$ and $c^{\text{tokens}}_{m, p}$ (\texttt{L16}), partially resynchronizing with the global trajectory without halting its progress. To determine when training is complete, the learner uses a global step count $t$ that is updated by the syncer (\texttt{L17}). This is necessary because learners may proceed at variable speeds, so by the time the desired number of global steps is reached, the learners will have completed different numbers of local steps.

\subsection{The Syncer: Global Aggregation}\label{sec:decoupled_syncer}

The syncer is responsible for reconciling the divergent states of the asynchronous learners via an outer optimization step. Crucially, this is where our approach diverges from standard data-parallel or synchronous DiLoCo methods: to guarantee high availability and maximize global goodput, the syncer does not wait for a synchronized consensus from all $M$ learners.

\begin{algorithm}[t!] \small
\caption{Decoupled DiLoCo: Learner} \label{alg:learner}
\begin{algorithmic}[1]
\Require Fragmented initial weights $\{\Theta_p^{(0)}\}_{p \in \mathcal{P}}$, Data shard $\mathcal{D}_m$
\Require Per-fragment sync interval $H$, Offsets $t_p$
\Require Optimizer \texttt{InnerOpt}
\State $\forall p: \theta_{m, p}^{(0)} \gets \Theta_p^{(0)}$
\State $\forall p: c^{\text{steps}}_{m, p} \gets 0, c^{\text{tokens}}_{m, p} \gets 0$
\State $t_m = 0$
\State $t = 0$
\While{\texttt{step} $t \le T$}
    \Statex
    \State \rlap{\Comment{\colorbox[HTML]{EEB211}{1. Inner optimization}}}
    \State $t_m = t_m + 1$
    \State $x \sim \mathcal{D}_m$
    \State $\forall p: c^{\text{steps}}_{m, p} \gets c^{\text{steps}}_{m, p} + 1,\; c^{\text{tokens}}_{m, p} \gets c^{\text{tokens}}_{m, p} + |x|$
    \State $\mathcal{L} \gets f(x, \theta_m^{(t_m-1)})$
    \State $\theta_m^{(t_m)} \gets \texttt{InnerOpt}(\theta_m^{(t_m-1)}, \nabla_\mathcal{L})$
    \Statex
    \State \rlap{\Comment{\colorbox[HTML]{EEB211}{2. Metadata learner $\rightarrow$ syncer}}}
    \State $\texttt{Send}(\text{Syncer}, t_m, \{c^{\text{steps}}_{m, p}\}_{p \in \mathcal{P}}, \{c^{\text{tokens}}_{m, p}\}_{p \in \mathcal{P}})$
    \Statex
    \State \rlap{\Comment{\colorbox[HTML]{EEB211}{3. Communication syncer $\rightarrow$ learner}}}
    \For{$(\Theta_p, t_s) \in \texttt{RecvAllPending}(\text{Syncer})$}
        \State $\theta_{m,p}^{(t)} \gets \Theta_p$
        \State $c^{\text{steps}}_{m, p} \gets 0,\; c^{\text{tokens}}_{m, p} \gets 0$
        \State $t = t_s$
    \EndFor
\EndWhile
\end{algorithmic}
\end{algorithm}

\begin{algorithm}[t!] \small
\caption{Decoupled DiLoCo: Syncer} \label{alg:syncer}
\begin{algorithmic}[1]
\Require Fragmented initial weights $\{\Theta_p^{(0)}\}_{p \in \mathcal{P}}$, $M$ learners
\Require Per-fragment sync interval $H$
\Require Minimum learners $K$
\Require Optimizer \texttt{OuterOpt}
\For{\texttt{step} $t = 1 \dots T$}
    \If{$\exists p$ s.t. $t \bmod H = t_p$}
    \Statex
    \State \rlap{\Comment{\colorbox[HTML]{EEB211}{1. Communication learner $\rightarrow$ syncer}}}
    \State $ \mathcal{U}_t \gets \texttt{Recv}(\text{Learners}, \text{at least } K) $
    \State $ \{ t_m, c^{\text{steps}}_{m,p}, c^{\text{tokens}}_{m,p} \}_{m \in \mathcal{M}_t} \gets \mathcal{U}_{t} $
    
    \For{$m \in \mathcal{M}_{t}$}
        \State $\theta^{(t)}_{m,p}=\texttt{Pull}(\text{Learner}_m, t_m, p)$
        \Statex
    \EndFor
    \State $w_{m,p} \gets \texttt{Weight}(\{c^{\text{steps}}_{m,p}, c^{\text{tokens}}_{m,p}\}_{m \in \mathcal{M}_t}) $

    \Statex
    \State \rlap{\Comment{\colorbox[HTML]{EEB211}{2. All-reduce across syncer shards}}}
    \State $\Delta_p^{(t)} \gets \mergefn\left(\{\theta_{m,p}^{(t)}, w_{m,p}\}_{m \in \mathcal{M}_{t}}, \Theta_{p}^{(t-H)}\right)$

    \Statex
    \State \rlap{\Comment{\colorbox[HTML]{EEB211}{3. Outer optimization}}}
    \State $\Theta_p^{(t)} \gets \texttt{OuterOpt}(\Theta_p^{(t-H)}, \Delta_p^{(t)})$

    \Statex
    \State \rlap{\Comment{\colorbox[HTML]{EEB211}{4. Communication syncer $\rightarrow$ learner}}}
    \State $\texttt{Send}(\text{Learners}, \Theta_p^{(t)}, t)$
    \EndIf
\EndFor
\end{algorithmic}
\end{algorithm}

We outline the syncer's operations in \autoref{alg:syncer}. The syncer maintains a global step count $t$. As in \eqref{eq:streaming_outer_opt}, when $t$ matches a predetermined offset $t_p \bmod H$ for fragment $p$ (\texttt{L2}), it will aggregate this fragment across learners and apply outer optimization. Rather than waiting for all $M$ learners, the syncer only waits for a minimum threshold $K \le M$ learners that successfully send their metadata (\texttt{L4}). Let $\mathcal{M}_t$ denote the subset of learners that successfully reported. The syncer then pulls the corresponding model fragment from all learners in $\mathcal{M}_t$. Any offline or straggling learners are excluded from the aggregation for that step, ensuring that localized hardware failures do not stall global training progress.

In a fully asynchronous system, fast learners naturally process more data between syncs than slow or recovering learners. To prevent the outer optimization from being disproportionately skewed by slower learners, the syncer calculates a dynamic weight $w_{m,p}$ for each learner using a weighting function $\texttt{Weight}$, derived from the reported per-learner $c^{\text{steps}}$ and $c^{\text{tokens}}$. Throughout, we use the following weighting function

\begin{align*}
    \texttt{Weight}(c^\text{tokens}_{m,p}, c^\text{steps}_{m,p}) = c^\text{tokens}_{m,p} \times \left( \frac{c^\text{tokens}_{m,p}}{c^\text{steps}_{m,p}} \right)\,.
\end{align*}

Intuitively, this corresponds to $w_{m, p}= \texttt{(quantity)} \times \texttt{(quality)}$, where a learner's contribution is of higher quality if it is amortized over fewer steps. The syncer then computes the outer gradient $\Delta_p^{(t)}$ using a merge function (\texttt{mergeFn}, \texttt{L11}). This function determines the aggregated shift by comparing the previous global state $\Theta_p^{(t-H)}$ against the weighted combination of the newly received learner fragments $\{\theta_{m,p}^{(t)}\}_{m \in \mathcal{M}_t}$. Finally, the syncer applies the outer optimizer (e.g., SGD with Nesterov momentum, \texttt{L13}) to derive an updated global fragment $\Theta_p^{(t)}$ which is asynchronously broadcasted back to the active learners (\texttt{L15}).

\paragraph{Adaptive Quorum Window.}\label{sec:adaptive_quorum}

When syncer operations outpace learner compute steps, overlapping communication and computation over $\tau$ steps creates a natural \emph{slack}. Instead of proceeding immediately once the minimum quorum $K$ is met, we use this idle time to introduce an adaptive \emph{grace window}, $\xi_{\text{grace}}$. This purposefully trades available network slack for improved sample efficiency by incorporating more learners into the global update, all without stalling the system or degrading goodput.

In our setting, setting $H=P$ and $\tau=2$ effectively double-buffers the training process. Our primary constraint is that gathering a quorum, the grace window, and synchronizing a fragment must fit within $\tau$ compute steps. The available slack is defined as:
\begin{equation}
    \xi_{\text{slack}} = \tau \times \xi_{\text{step}} - (\xi_{\text{quorum}} + \xi_{\text{sync}})\,
\end{equation}
where $\xi_{\text{step}}$, $\xi_{\text{quorum}}$, and $\xi_{\text{sync}}$ are the durations for a compute step, reaching quorum, and synchronization, respectively. To prevent bottlenecks, the grace window $ \xi_{\text{grace}} \le \gamma \cdot \xi_{\text{slack}}$ is bounded by a safety margin $\gamma < 1$. In practice, $\xi_{\text{step}}$ and $\xi_{\text{quorum}}$ can be tracked via exponential moving averages to adapt to changing speeds and bandwidth.

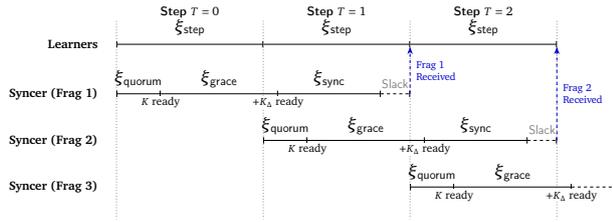
\begin{figure}[htbp]
    \centering
    \resizebox{\columnwidth}{!}{%
        \begin{tikzpicture}[
            font=\normalsize\sffamily,
            >=Stealth,
            every node/.style={align=center}
        ]
        
        \def\stepw{5}      
        \def\qwait{1.5}    
        \def\gwait{4}      
        \def\sync{3.5}     
        
        \pgfmathsetmacro{\gw}{\qwait+\gwait}
        \pgfmathsetmacro{\sw}{\gw+\sync}
        \pgfmathsetmacro{\endI}{2*\stepw}
        
        \pgfmathsetmacro{\qwII}{\stepw+\qwait}
        \pgfmathsetmacro{\gwII}{\stepw+\gw}
        \pgfmathsetmacro{\swII}{\stepw+\sw}
        \pgfmathsetmacro{\endII}{3*\stepw}

        \pgfmathsetmacro{\qwIII}{2*\stepw+\qwait}
        \pgfmathsetmacro{\gwIII}{2*\stepw+\gw}
        
        \def\ycomp{2.5}    
        \def\yfI{0.8}      
        \def\yfII{-0.8}    
        \def\yfIII{-2.4}   
        
        \draw[thick, |-|] (0, \ycomp) -- (\stepw, \ycomp) node[midway, above=4pt] {\large\textbf{Step $T=0$} \\ \LARGE $\xi_{\text{step}}$};
        \draw[thick, -|] (\stepw, \ycomp) -- (2*\stepw, \ycomp) node[midway, above=4pt] {\large\textbf{Step $T=1$} \\ \LARGE $\xi_{\text{step}}$};
        \draw[thick, -|] (2*\stepw, \ycomp) -- (3*\stepw, \ycomp) node[midway, above=4pt] {\large\textbf{Step $T=2$} \\ \LARGE $\xi_{\text{step}}$};
        
        \draw[dotted, gray, thick] (0, \ycomp+0.8) -- (0, -3.5);
        \draw[dotted, gray, thick] (\stepw, \ycomp+0.8) -- (\stepw, -3.5);
        \draw[dotted, gray, thick] (2*\stepw, \ycomp+0.8) -- (2*\stepw, -3.5);
        \draw[dotted, gray, thick] (3*\stepw, \ycomp+0.8) -- (3*\stepw, -3.5);
        
        \draw[thick, |-|] (0, \yfI) -- (\qwait, \yfI) node[midway, above=2pt] {\LARGE $\xi_{\text{quorum}}$};
        \node[below=2pt, font=\normalsize, fill=white, inner sep=1pt] at (\qwait, \yfI) {$K$ ready};
        
        \draw[thick, -|] (\qwait, \yfI) -- (\gw, \yfI) node[midway, above=2pt] {\LARGE $\xi_{\text{grace}}$};
        \node[below=2pt, font=\normalsize, fill=white, inner sep=1pt] at (\gw, \yfI) {$+K_\Delta$ ready};
        
        \draw[thick, -|] (\gw, \yfI) -- (\sw, \yfI) node[midway, above=2pt] {\LARGE $\xi_{\text{sync}}$};
        \draw[thick, dashed, -|] (\sw, \yfI) -- (\endI, \yfI) node[midway, above=2pt, text=gray] {\large Slack};
        
        \draw[->, thick, blue!80!black, dashed] (\endI, \yfI) -- (\endI, \ycomp-0.1) node[midway, right=2pt, align=left, font=\small] {Frag 1 \\ Received};
        
        \draw[thick, |-|] (\stepw, \yfII) -- (\qwII, \yfII) node[midway, above=2pt] {\LARGE $\xi_{\text{quorum}}$};
        \node[below=2pt, font=\normalsize, fill=white, inner sep=1pt] at (\qwII, \yfII) {$K$ ready};
        
        \draw[thick, -|] (\qwII, \yfII) -- (\gwII, \yfII) node[midway, above=2pt] {\LARGE $\xi_{\text{grace}}$};
        \node[below=2pt, font=\normalsize, fill=white, inner sep=1pt] at (\gwII, \yfII) {$+K_\Delta$ ready};
        
        \draw[thick, -|] (\gwII, \yfII) -- (\swII, \yfII) node[midway, above=2pt] {\LARGE $\xi_{\text{sync}}$};
        \draw[thick, dashed, -|] (\swII, \yfII) -- (\endII, \yfII) node[midway, above=2pt, text=gray] {\large Slack};
        
        \draw[->, thick, blue!80!black, dashed] (\endII, \yfII) -- (\endII, \ycomp-0.1) node[midway, right=2pt, align=left, font=\small] {Frag 2 \\ Received};
        
        \draw[thick, |-|] (2*\stepw, \yfIII) -- (\qwIII, \yfIII) node[midway, above=2pt] {\LARGE $\xi_{\text{quorum}}$};
        \node[below=2pt, font=\normalsize, fill=white, inner sep=1pt] at (\qwIII, \yfIII) {$K$ ready};
        
        \draw[thick, -|] (\qwIII, \yfIII) -- (\gwIII, \yfIII) node[midway, above=2pt] {\LARGE $\xi_{\text{grace}}$};
        \node[below=2pt, font=\normalsize, fill=white, inner sep=1pt] at (\gwIII, \yfIII) {$+K_\Delta$ ready};
        \draw[thick, dashed, -] (\gwIII, \yfIII) -- (\gwIII+1.5, \yfIII);
        
        \node[anchor=east, font=\large\bfseries] at (-0.5, \ycomp) {Learners};
        \node[anchor=east, font=\large\bfseries] at (-0.5, \yfI) {Syncer (Frag 1)};
        \node[anchor=east, font=\large\bfseries] at (-0.5, \yfII) {Syncer (Frag 2)};
        \node[anchor=east, font=\large\bfseries] at (-0.5, \yfIII) {Syncer (Frag 3)};
        
        \end{tikzpicture}%
    }
    \caption{Illustration of the adaptive quorum grace window fitting within the available slack.}
    \label{fig:grace_window}
\end{figure}

Illustrated in \autoref{fig:grace_window}, this brief wait trades excess bandwidth for sample efficiency by incorporating more learners. Combined with token-based weighting $w_{m,p}$, slightly delayed updates can still contribute to the global update, reducing outer gradient variance while naturally handling hardware speed discrepancies. We also found that in practice, when learner steps vary in speed or are not aligned, an adaptive grace window works better than setting higher $K$ for maintaining a consistent sync frequency without introducing gaps in utilization.

\subsection{Chaos engineering for LLMs}\label{sec:simulating_failures}

\begin{table*}[htbp]
\definecolor{diag0}{HTML}{D73027} 
\definecolor{diag1}{HTML}{F46D43} 
\definecolor{diag2}{HTML}{FDAE61} 
\definecolor{diag3}{HTML}{FEE08B} 
\definecolor{diag4}{HTML}{FFFFBF} 
\definecolor{diag5}{HTML}{D9EF8B} 
\definecolor{diag6}{HTML}{A6D96A} 
\definecolor{diag7}{HTML}{66BD63} 
\definecolor{diag8}{HTML}{1A9850} 
\definecolor{graycell}{HTML}{D3D3D3} 

\centering

\begin{subtable}[t]{0.48\linewidth}
\vspace{0pt}
\centering
\resizebox{\columnwidth}{!}{
\begin{tabular}{@{}r r r r r r@{}}
\toprule
\textbf{Goodput} & \multicolumn{5}{c}{Number of chips $N_\text{chip}$} \\
\cmidrule(l){2-6}
\# Learners $M$ & \textbf{150k} & \textbf{300k} & \textbf{600k} & \textbf{1.2m} & \textbf{2.4m} \\
\midrule
(no elasticity) \textbf{1} & \cellcolor{graycell} 72\% & \cellcolor{graycell} 57\% & \cellcolor{graycell} 38\% & \cellcolor{graycell} 27\% & \cellcolor{graycell} 18\%
\vspace{0.2cm} \\
\textbf{1}  & \cellcolor{diag4} 88\% & \cellcolor{diag3} 80\% & \cellcolor{diag2} 69\% & \cellcolor{diag1} 58\% & \cellcolor{diag0}\textcolor{white}{40\%} \\
\textbf{2}  & \cellcolor{diag5} 94\% & \cellcolor{diag4} 88\% & \cellcolor{diag3} 80\% & \cellcolor{diag2} 70\% & \cellcolor{diag1} 54\% \\
\textbf{4}  & \cellcolor{diag6} 96\% & \cellcolor{diag5} 93\% & \cellcolor{diag4} 89\% & \cellcolor{diag3} 82\% & \cellcolor{diag2} 73\% \\
\textbf{8}  & \cellcolor{diag7} 98\% & \cellcolor{diag6} 96\% & \cellcolor{diag5} 94\% & \cellcolor{diag4} 88\% & \cellcolor{diag3} 80\% \\
\textbf{16} & \cellcolor{diag8}\textcolor{white}{99\%} & \cellcolor{diag7} 98\% & \cellcolor{diag6} 96\% & \cellcolor{diag5} 93\% & \cellcolor{diag4} 86\% \\
\bottomrule
\end{tabular}
}
\caption{\footnotesize \textbf{Goodput}: \% of the cluster doing useful work.}
\label{subtab:goodput}
\end{subtable}%
\hfill 
\begin{subtable}[t]{0.48\linewidth}
\vspace{0pt}
\centering
\resizebox{\columnwidth}{!}{
\begin{tabular}{@{}r r r r r r@{}}
\toprule
\textbf{System Uptime} & \multicolumn{5}{c}{Number of chips $N_\text{chip}$} \\
\cmidrule(l){2-6}
\# Learners $M$ & \textbf{150k} & \textbf{300k} & \textbf{600k} & \textbf{1.2m} & \textbf{2.4m} \\
\midrule
(no elasticity) \textbf{1} & \cellcolor{graycell} 72\% & \cellcolor{graycell} 57\% & \cellcolor{graycell} 38\% & \cellcolor{graycell} 27\% & \cellcolor{graycell} 18\%
\vspace{0.2cm}\\
\textbf{1}  & \cellcolor{diag4} 88\% & \cellcolor{diag3} 81\% & \cellcolor{diag2} 69\% & \cellcolor{diag1} 59\% & \cellcolor{diag0}\textcolor{white}{42\%} \\
\textbf{2}  & \cellcolor{diag8}\textcolor{white}{100\%} & \cellcolor{diag7} 97\% & \cellcolor{diag6} 96\% & \cellcolor{diag5} 90\% & \cellcolor{diag3} 80\% \\
\textbf{4}  & \cellcolor{diag8}\textcolor{white}{100\%} & \cellcolor{diag8}\textcolor{white}{100\%} & \cellcolor{diag7} 99\% & \cellcolor{diag7} 99\% & \cellcolor{diag7} 99\% \\
\textbf{8}  & \cellcolor{diag8}\textcolor{white}{100\%} & \cellcolor{diag8}\textcolor{white}{100\%} & \cellcolor{diag8}\textcolor{white}{100\%} & \cellcolor{diag8}\textcolor{white}{100\%} & \cellcolor{diag7} 99\% \\
\textbf{16} & \cellcolor{diag8}\textcolor{white}{100\%} & \cellcolor{diag8}\textcolor{white}{100\%} & \cellcolor{diag8}\textcolor{white}{100\%} & \cellcolor{diag8}\textcolor{white}{100\%} & \cellcolor{diag7} 99\% \\
\bottomrule
\end{tabular}
}
\caption{\footnotesize \textbf{System Uptime}: \% of the time a cluster is stepping.}
\label{subtab:uptime}
\end{subtable}

\vspace{0.cm} 
\caption{\textbf{Modeled goodput and system uptime for hypothetical configurations} across number of decoupled learners $M$ and number of simulated chips $N_\text{chip}$ for a fixed $\texttt{MTBI}_\text{chip}=1\,\text{year}$ level of simulated hardware failures. Here we assume all chips have constant speed. Note that $M=1$ learner is equivalent to data-parallel. }
\label{tab:goodput_across_m_and_c}
\end{table*}

Chaos engineering \citep{basiri2016chaos} tests system resilience by simulating failures, a practice originally designed for general infrastructure. We apply this principle to LLM training, anchoring in the observation that determinism and reproducibility are critical for enhancing system resilience, debugging, and algorithmic scaling properties.

We distill the wide range of possible hardware failures during LLM training \citep{LlamaTeam2024} into five key parameters: the mean time between interruptions (\texttt{MTBI}) per chip ($\texttt{MTBI}_{\text{chip}}$), the total number of simulated chips ($N_\text{chip}$), the processing speed variance per chip, the duration required for elastic downscaling and upscaling, and the time needed for a failed chip to return online. Crucially, for a fixed interruption rate per chip, the mean time between failures (MTBF) of the entire cluster decreases proportionally as the number of chips increases:
\begin{equation}
    \texttt{MTBF}_{\text{cluster}} = \frac{\texttt{MTBI}_{\text{chip}}}{N_\text{chip}}\,.
\end{equation}

We experiment with strategies for resilience by generating tapes of events that would be logged during a system execution that encountered the failures of interest (detailed in Section~\ref{sec:tape_generation}). The number of chips failing at any given step is sampled from a Poisson distribution parameterized by the inverse of $\text{MTBF}_{\text{cluster}}$. When a chip fails, its entire slice is temporarily removed and only returns after a delay sampled from an exponentiated Weibull distribution. The subsequent elastic downscaling and upscaling operations consume a user-configurable, fixed amount of time.

This simulation yields a \textit{goodput} metric \citep{wongpanich2025machinelearningfleetefficiency}: the percentage of allocated cluster time actually spent executing useful steps; a training run that executes some of the steps with less slices available than expected will have lower goodput. While we omit Model Flops Utilization (MFU) for simplicity, our decoupled framework introduces no MFU regression.

\autoref{tab:goodput_across_m_and_c} reports goodput across $M \in \{1,2,4,8,16\}$ learners (where $M=1$ is standard data parallelism) under various amount of simulated chips. Assuming an $\texttt{MTBI}_{\text{chip}}$ of 1 year and a number of chips $N_\text{chip}$ chosen arbitrarily for this paper and reconfiguration times of tens of seconds, standard data-parallel training without elasticity responds poorly to failures. Even with elasticity, goodput plummets to 40\% for $N_\text{chip}=2.4\text{m}$ chips. Conversely, decoupling training across multiple learners ($M > 1$) maintains consistently high goodput. We also report in \autoref{subtab:uptime}, the system uptime (e.g. how often the system is stepping): notably, with sufficiently large $M$, the uptime can be 100\%, with not a single downtime.

\begin{tcolorbox}[colback=boxblue, colframe=black, arc=4pt, boxsep=0.3pt]
\textbf{Goodput Rule of Thumb:} An $M$-learner system matches the goodput of a single-learner system using $M$ times fewer chips.

{\small $\texttt{goodput}(M, N_\text{chip}) \approx \texttt{goodput}(k \times M, k \times N_\text{chip})$}

\end{tcolorbox}

For simulation purposes, we simulate a significantly larger number of chips ($\mathcal{O}(1\text{m})$) than we physically deploy ($\mathcal{O}(1\text{k})$). When a simulated chip fails, its corresponding slice is discarded, reducing the effective batch size, similar to the slice-granularity elasticity described by \citet{geminiteam2025gemini2p5}. During our ML experiments, we proportionally scale down our real batch size to match what would occur at the larger simulated scale.

\section{System design}\label{sec:infrastructure}

\begin{figure*}[ht]
    \centering
    \includegraphics[width=\textwidth, trim={0cm 7cm 6cm 0cm}, clip]{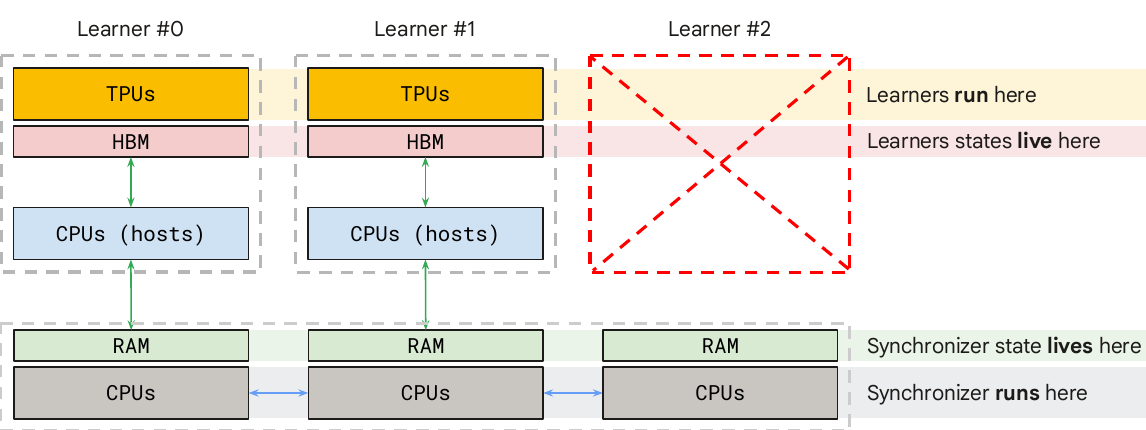}
    \caption{\textbf{Overview of the Decoupled DiLoCo system architecture}. Each learner worker runs an independent data-parallel training loop on a partition of the accelerator mesh. Learner workers communicate with the syncer by sending parameter fragments and receiving outer-optimized updates over the data-center network (DCN). The syncer is $M$-way sharded across CPU-only replicas and performs the outer optimization step. Learner workers may be temporarily absent (here learner \#2), but their corresponding syncer shards persist, enabling dynamic adjustment of the number of active learners. After each step, the updated learner model is copied to RAM of the host CPUs colocated with learner TPUs. This allows the syncer to select the fragment to transmit over the data-center network while the learner continues performing inner optimization steps, without requiring extra model copies on HBM.}
    \vspace{-0.5cm}
    \label{fig:system}
\end{figure*}

Current model pretraining relies on a globally synchronized execution model. While devices continuously execute computation and communication primitives, they are bound by strict synchronization barriers: a failure or slowdown in any single device stalls the entire cluster. The outermost parallelism level in standard pretraining is distributed data parallelism (DP), where multiple model replicas compute gradients with respect to the shared parameters, which are then averaged via collective communication. A critical constraint of this paradigm is that all replicas must succeed and participate on every step. Our system design removes this tight coupling by allowing replicas to take steps independently of one another. This naturally leads to a parameter server architecture, inspired by \citep{dean2012large}.

All workers are orchestrated by Pathways~\citep{barham2022pathways}, which manages the resource allocation, device mesh construction, and inter-worker dataflow. Each of the workers described in the following section is driven by a separate Pathways client, in order to prevent a single multi-threaded client from becoming a bottleneck.

\paragraph{Learner workers.}

The system consists of $M$ \emph{learner workers}, each of which runs an independent training loop. A learner worker can be viewed as a scaled-down version of a conventional DP job: if the full cluster provides capacity for $R$ data-parallel replicas, each learner worker operates with $R / M$ replicas. Each learner worker independently compiles and initializes its model, loads training data from its assigned shard, and executes inner optimization steps (e.g., AdamW) without coordination with the other learners. Crucially, the learner workers are \emph{isolated from one another}---they share no accelerator resources and do not communicate directly except in the special case of recovery, the process by which one learner obtains up-to-date state from another learner in order to rejoin after a long failure (see Section~\ref{sec:recovery}). This isolation is the foundation for both decoupling and failure domain containment: a hardware failure or straggler machine in one learner has no effect on the others.

\paragraph{Syncer worker.}
The learner workers are connected to a syncer worker, which instantiates the role described in Section~\ref{sec:decoupled_syncer}. The syncer worker maintains the global model parameters and outer optimizer state, and is responsible for aggregating the learner updates and broadcasting the outer-optimized parameters back to the learners. Since it only holds parameters and optimizer state (no activations) and performs relatively simple element-wise operations, its computational and memory footprint is significantly smaller than that of any learner worker. Accordingly, the syncer worker runs on CPU-only resources, partitioned into $M$ shards---one corresponding to each learner worker.

As illustrated in \autoref{fig:system}, a learner worker may be temporarily absent (e.g., due to a hardware failure), yet its corresponding syncer shard persists. Thus, when a learner is unavailable, the syncer worker can execute the same \texttt{Merge} computation as usual, with that learner's weight set to 0. This design enables the system to dynamically adjust the number of active learners without modifying the syncer worker's configuration or requiring a restart.

\paragraph{State coordination.}
Each worker locally maintains a vector clock \citep{mattern1989virtual} keeping track of its own step as well as its latest knowledge of the step for each of the other workers in the system. Learner and syncer workers communicate training progress via message passing over FIFO channels where the sender attaches its current vector clock to each message, as shown in \autoref{fig:checkpointing}. These vector clocks are the basis for many operations critical to reliable system operation, including watermarking of channels to garbage-collect old state, creating consistent global checkpoints via the Chandy-Lamport distributed snapshotting algorithm \citep{chandy1985distributed} as discussed in Section~\ref{sec:checkpointing}, and logging of a nondeterministic training run to enable deterministic replay as discussed in Section~\ref{sec:tape_generation}.

\paragraph{Summary of benefits.}

This architecture yields several advantages over monolithic data-parallel training. First, it ensures a \textbf{reduced synchronous footprint with isolated failure domains}: by partitioning the cluster into $M$ smaller accelerator groups, the \texttt{MTBF} of each group improves proportionally (see \autoref{tab:goodput_across_m_and_c}), and the blast radius of any hardware failure is strictly contained, preventing it from propagating to the rest of the system. Second, it enables \textbf{resilient coordination} by running the syncer worker on highly stable, CPU-only resources with a minimal failure surface. While not the primary objective of this design, the system inherently retains the \textbf{massive bandwidth reduction} properties of its predecessor, Streaming DiLoCo \citep{douillard2025streaming}, due to the fact that learners and syncer exchange \emph{parameter fragments} over the data-center network, and at each outer optimization step the syncer shards perform an all-reduce over only a single fragment rather than the whole model (see Section~\ref{sec:bandwidth_profile}). Finally, training across \textbf{heterogeneous resources} (e.g. chips of different generations) becomes straightforward in this setting: the decoupled learners are free to choose their own hardware, and the loose nature of the synchronization mitigates speed differences inherent from attempts to balance workload across heterogeneous compute. A combination of all those advantages also enables us to \textit{scavenge} pre-emptible heterogeneous compute resources distributed across far apart locations.

\begin{figure*}[ht]
    \centering
    \includegraphics[width=1.0\textwidth]{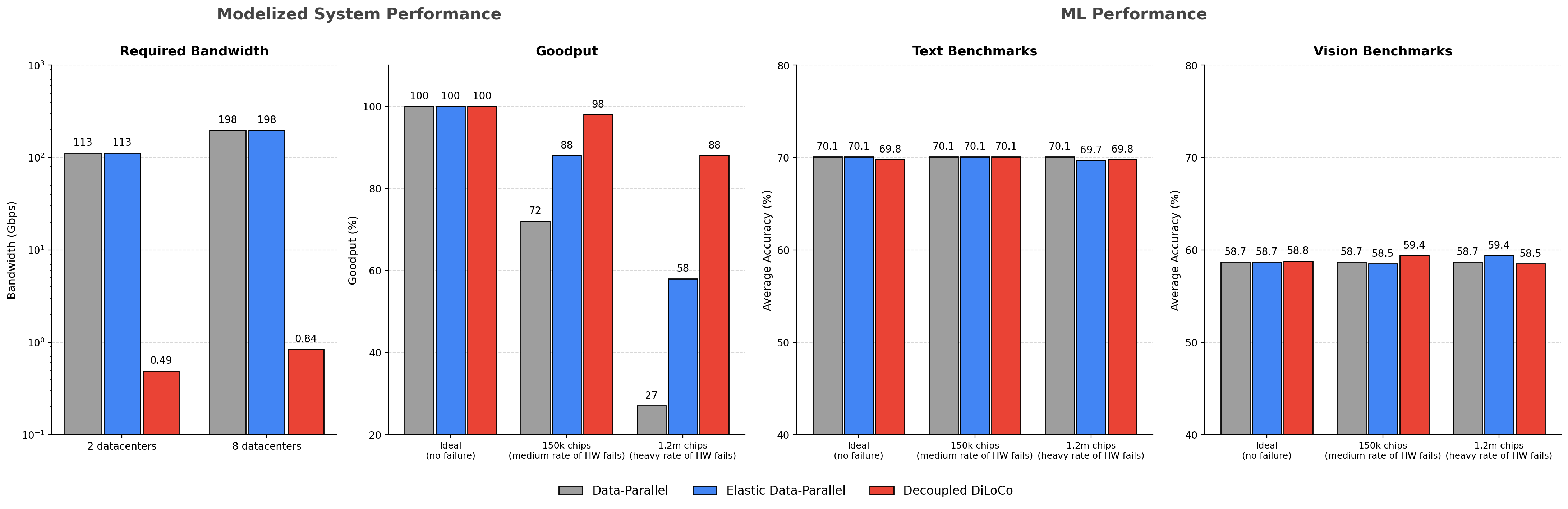}
    \caption{\textbf{Hardware failures resilience} of Decoupled DiLoCo vs elastic data-parallel with a dense 5B model trained on 1T tokens. Compared to elastic data-parallel, Decoupled DiLoCo requires two orders of magnitude less bandwidth, observes significantly superior goodput under a heavy rate of hardware failures, and yet performs equivalently on pure ML performance.}
    \label{fig:async_hw}
\end{figure*}

\begin{table}[ht]
\centering
\resizebox{\columnwidth}{!}{
\setlength{\tabcolsep}{2.5pt} 
\begin{tabular}{l ccc ccc}
\toprule
\multirow{2}{*}{} & \multicolumn{3}{c}{Elastic DP} & \multicolumn{3}{c}{Decoupled $M=8$} \\
\cmidrule(lr){2-4} \cmidrule(l){5-7}
\# chips & -- & 150k & 1.2m & -- & 150k & 1.2m\\
$\texttt{MTBI}_\text{chip}$ & $+\infty$ & 1y & 1y & $+\infty$ & 1y & 1y \\
\midrule
Goodput        & \textbf{100\%} & 88\% & 58\% & \textbf{100\%} & \textbf{98\%} & \textbf{88\%} \\
\midrule
\textcolor{gray}{\textit{Text (Avg)}} & \textcolor{gray}{70.1} & \textcolor{gray}{70.1} & \textcolor{gray}{69.7} & \textcolor{gray}{69.8} & \textcolor{gray}{70.1} & \textcolor{gray}{69.8} \\
Arc-Chall.     & \textbf{61.2} & 60.4 & 60.4 & 59.0 & 60.3 & 59.2 \\
Arc-Easy       & \textbf{81.9} & 81.2 & 80.9 & 81.2 & 81.4 & 81.6 \\
BoolQ          & 76.8 & 78.7 & 78.4 & 76.8 & \textbf{79.4} & 78.1 \\
HellaSwag      & 74.4 & 74.40 & 73.7 & 74.2 & \textbf{74.5} & 74.2 \\
Piqa           & 79.8 & 79.5 & 79.2 & \textbf{79.9} & 79.5 & 79.1 \\
Siqa           & 50.3 & \textbf{50.7} & 49.6 & 49.9 & 49.9 & 49.5 \\
Winogrande     & 66.6 & 65.6 & 65.5 & \textbf{67.7} & 65.6 & 67.2 \\
\midrule
\textcolor{gray}{\textit{Vision (Avg)}} & \textcolor{gray}{58.7} & \textcolor{gray}{58.5} & \textcolor{gray}{59.4} & \textcolor{gray}{58.8} & \textcolor{gray}{59.4} & \textcolor{gray}{58.6} \\
MMMU           & 33.9 & 32.4 & \textbf{36.2} & 33.4 & 32.4 & 33.2 \\
COCO-Cap.      & 90.1 & 89.9 & \textbf{92.2} & 88.2 & 90.8 & 89.5 \\
ChartQA        & 25.1 & 25.0 & 26.6 & 24.8 & 24.8 & \textbf{27.4} \\
DocVQA         & 79.6 & 78.3 & 77.8 & \textbf{80.9} & 80.0 & 76.7 \\
DocVQA-Hard    & 56.6 & 56.9 & 56.9 & 59.7 & \textbf{59.8} & 55.7 \\
InfographicVQA & 48.3 & 48.5 & 47.7 & 48.1 & \textbf{50.4} & 48.9 \\
TextVQA        & 77.0 & 78.4 & 78.3 & 76.7 & 77.6 & \textbf{78.4} \\
\bottomrule
\end{tabular}
}
\caption{\textbf{Simulated hardware failure impact} on ML performance of a 5B dense with 1T tokens. Given a fixed MTBI per chip, we increase the number of simulated chips, lowering the cluster MTBF and leading to more frequent hardware failures. Decoupled DiLoCo’s ML performance is not degraded by the algorithmic changes to achieve  significantly higher goodput.}
\label{tab:tab_hw_failures}
\end{table}

\begin{table}[h]
\centering
\resizebox{0.8\columnwidth}{!}{
\setlength{\tabcolsep}{2.5pt} 
\begin{tabular}{l c cc}
\toprule
\multirow{2}{*}{} & \multicolumn{1}{c}{Elastic DP} & \multicolumn{2}{c}{Decoupled $M=8$} \\
\cmidrule(lr){2-2} \cmidrule(l){3-4}
\# chips & -- & -- & 1.2m \\
$\texttt{MTBI}_\text{chip}$ & $+\infty$ & $+\infty$ & 1y \\
\midrule
Goodput        & \textbf{100\%} & \textbf{100\%} & \textbf{88\%} \\
\midrule
\textcolor{gray}{\textit{Text (Avg)}} & \textcolor{gray}{68.4} & \textcolor{gray}{67.9} & \textcolor{gray}{68.4} \\
Arc-Chall.     & 59.1 & 55.5 & \textbf{59.4} \\
Arc-Easy       & \textbf{81.7} & 80.6 & 81.4 \\
BoolQ          & 71.1 & \textbf{72.5} & 71.8 \\
HellaSwag      & \textbf{73.5} & 73.4 & 73.1 \\
Piqa           & \textbf{79.8} & 79.3 & \textbf{79.8} \\
Siqa           & 49.1 & \textbf{50.3} & 49.3 \\
Winogrande     & \textbf{64.6} & 63.7 & 64.2 \\
\midrule
\textcolor{gray}{\textit{Vision (Avg)}} & \textcolor{gray}{54.9} & \textcolor{gray}{55.1} & \textcolor{gray}{54.3} \\
MMMU           & 30.4 & 29.8 & \textbf{31.1} \\
COCO-Cap.      & \textbf{89.0} & 84.5 & 84.9 \\
ChartQA        & 20.9 & \textbf{21.9} & 20.6 \\
DocVQA         & 74.7 & \textbf{76.9} & 74.0 \\
DocVQA-Hard    & 51.1 & \textbf{51.6} & 51.3 \\
InfographicVQA & 43.4 & \textbf{46.9} & 43.1 \\
TextVQA        & 74.8 & 74.2 & \textbf{75.0} \\
\bottomrule
\end{tabular}
}
\caption{\textbf{Simulated hardware failures impact} on ML performance of a 2.8B activated Mixture-of-Experts model trained on 170B tokens, using the same setup as in \autoref{tab:tab_hw_failures}.}
\label{tab:moe_tab_hw_failures}
\end{table}

\section{Experiments}\label{sec:experiments}

We now detail the experiments we conducted to validate our framework. For all experiments, we use Gemma 4 \citep{gemma4_2026} customized for a lighter training footprint, and train on a mixture of text and vision data.

\subsection{Experimental Setup}

In all experiments, we use $P=24$ fragments. Each fragment is synchronized once every $H=24$ steps. The synchronization is overlapped over $\tau=2$ steps. As a consequence, at every step a fragment is sent and received, and there are two fragments in flight. Unless stated otherwise, we use the minimal quorum size $K=1$ and use a grace window (see Section~\ref{sec:adaptive_quorum}) that can extend up to a step.

We form the fragments via a greedy bin-packing algorithm, applied to individual tensors in the model. This results in approximately balanced fragments. We found that this strategy, which we refer to as \emph{balanced tensor fragmentation}, maintained model quality compared to layer-based fragmentation strategies~\citep{douillard2025streaming}, while significantly reducing peak bandwidth. See Section~\ref{sec:fragmentation_ablations} for details.

We do not just use direct averaging for the \mergefn operation in \autoref{alg:syncer}. We propose a modified merging operation, \emph{Radial-Directional Averaging} (RDA) in which we separately average the norms and directions of outer gradients. This leads to greater hyperparameter stability in the outer optimizer and boosts performance when scaling the number of learners $M$. For details, see Section~\ref{sec:merging_ablations}.

\subsection{Resilience to hardware failures}\label{sec:exp_hardware_failures}

In order to test the ML performance of Decoupled DiLoCo under heavy rates of failure, we simulate different scenarios. We choose to set an aggressive $\texttt{MTBI}_{\text{chip}} = 1\, \text{year}$ and we vary the amount of chips from $N_\text{chip}=150\text{k}$ ($1\times)$ to $N_\text{chip}=1.2\text{m}$ ($8\times$). The latter leads to a $\texttt{MTBF}_\text{cluster}$ of less than a minute. We display in \autoref{fig:async_hw} the downstream performance of both Data-Parallel (DP) and Decoupled DiLoCo $M=8$ on text and vision tasks alongside the observed goodput at various hardware failures rate. We also detailed the individual evaluation results in \autoref{tab:tab_hw_failures}.

Note that in the elastic data-parallel baseline, ML performance remains invariant to the hardware failure rate, but goodput degrades drastically, resulting in significantly longer wall-clock training times. The impact of hardware failures on training dynamics is twofold: (1) for both data-parallel and Decoupled DiLoCo, the loss of chips and their corresponding slices reduces the effective batch size, and (2) specific to Decoupled DiLoCo, recovering learners rejoin the system with stale parameters and optimizer states.

Notice that with $M=8$, the goodput degrades gracefully as the number of chips is increased (and thus the cluster $\texttt{MTBF}$ is lowered) up to 88\% while data-parallel, even with elasticity, reaches 58\% goodput. The downstream evaluations on both text and vision remain competitive, showing both the resiliency to hardware failures and ML robustness of our framework. 

We also validate that the downstream performance, in the event of frequent hardware failures, has similar impacts on mixture-of-experts models. In \autoref{tab:moe_tab_hw_failures}, we compare downstream performance of Data-Parallel and Decoupled DiLoCo $M=8$ with no hardware failures to that of Decoupled DiLoCo with $\texttt{MTBI}_{\text{chip}} = 1\, \text{year}$ and $N_\text{chip}=1.2\text{m}$ ($8\times$). We find that the evaluation results are all essentially comparable, with minor differences likely attributable to an ambient noise floor versus any algorithmic or system issue.

\FloatBarrier

\begin{table}[h]
\centering
\resizebox{\columnwidth}{!}{
\setlength{\tabcolsep}{4pt}

\begin{tabular}{ll ccc}
\toprule
Model & Scavenging & Text (Avg) & Vision (Avg) & Train Time \\
\midrule
\multirow{5}{*}{DP} & $+0\%$ & $60.1$ & $46.4$ & $1.00\times$ \\
& $+25\%$ & $60.5$ & $46.6$ & $1.07\times$ \\
& $+50\%$ & $60.0$ & $45.9$ & $1.09\times$ \\
& $+100\%$ & $60.1$ & $47.0$ & $1.02\times$ \\
& $+300\%$ & $60.4$ & $46.4$ & $0.80\times$ \\
\cmidrule(lr){1-5}
\multirow{5}{*}{DiLoCo} & $+0\%$ & $60.6$ & $46.2$ & $1.00\times$ \\
& $+25\%$ & $60.8$ & $\mathbf{47.2}$ & $0.90\times$ \\
& $+50\%$ & $\mathbf{61.3}$ & $45.2$ & $0.83\times$ \\
& $+100\%$ & $60.7$ & $46.4$ & $0.75\times$ \\
& $+300\%$ & $60.7$ & $46.1$ & $\mathbf{0.62\times}$ \\
\bottomrule
\end{tabular}
}
\caption{\textbf{Iso-FLOPs scavenging} experiments with a 2B dense trained on 128B tokens as more TPUs are scavenged across distributed locations for Decoupled DiLoCo and data-parallel. All runs share the same flops \& token budget, thus scavenging allows finishing earlier. Scavenged compute is available for half the FLOPS in each run, so the lower bound on relative training time is $0.5\times$.}
\label{tab:isoflop_avg_comparison}
\end{table}

\begin{figure*}[h]
    \centering
    \includegraphics[width=1.0\linewidth]{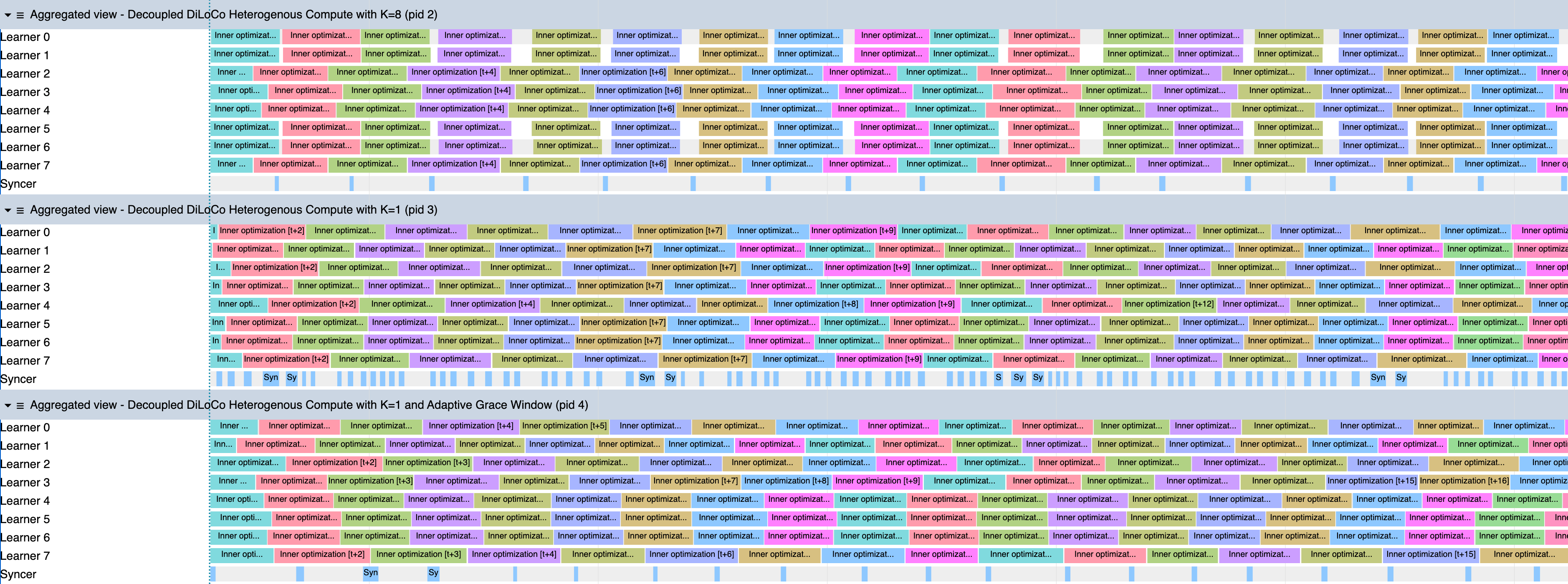}
    \caption{\textbf{XLA operations} when learners have varying step times for quorum sizes of $K=8$, $K=1$, and $K=1$ with an adaptive grace window. Notice that despite variable bandwidth and hardware heterogeneity there is always a learner stepping, achieving maximum availability.}
    \label{fig:xprof_heterogeneous_speed}
\end{figure*}

\subsection{Scavenging}\label{sec:exp_scavenging}

We exploit our decoupled framework to dynamically add compute resources, ``\emph{scavenge}'', on the fly during training. While slice-granular elasticity \citep{geminiteam2025gemini2p5} allows for scaling up the slice count, it incurs a non-negligible goodput penalty during system reconfiguration. Decoupled DiLoCo, however, seamlessly integrates any temporarily available extra FLOPS by dynamically acquiring (and releasing) extra learners using the learner recovery procedure described in Section~\ref{sec:recovery} .

For the duration that more learners are available the system runs with a temporarily higher $M$ than baseline training. In these experiments we use $M = 4$ learners as a base level of always available compute, and increase the number of learners to $M=5$, $6$, $8$ or $16$ intermittently during training to simulate a day-night cycle of increased availability. For a point of comparison we simulate Data-Parallel training being used in identical scavenging patterns, by increasing the batch size during periods of high $M$ to match Decoupled DiLoCo's step-by-step compute and token usage. This gives a strong baseline for an ML performance target but incurs the practical downsides of using distributed DP training for scavenging, such as the time taken to transfer the current model state to the new learners during batch upsize, and the increased step time when all-reducing gradients over increasingly geo-separated compute that has no pre-allocated bandwidth due to the opportunistic nature of scavenging.

We run our experiments in an \emph{iso-FLOPs} regime where total compute usage is fixed; the goal of this style of scavenging is therefore to accelerate training by reducing steps taken and wall-clock time. Table~\ref{tab:isoflop_avg_comparison} shows Decoupled DiLoCo is able to reap the benefits of ad hoc training acceleration without degrading ML performance, and does so more time-efficiently than the DP baseline which needs to more than double the number of learners to see any benefit. Section~\ref{sec:appendix_scavenging_results} contains a full breakdown of the downstream performance metrics.

\subsection{Heterogeneous hardware}\label{sec:exp_hetero_hw}

\begin{table}[!b]
\centering
\small 
\setlength{\tabcolsep}{6pt} 
\resizebox{\columnwidth}{!}{
\begin{tabular}{l ccc}
\toprule
\cmidrule(l){2-4} 
Quorum ($K$) & 8 & 1 & 1 \\
Grace Window ($\xi_{\text{grace}}$) & Off & Off & On \\
\midrule
Goodput        & 84\% & \textbf{100\%} & \textbf{100\%} \\
\midrule
\textcolor{gray}{\textit{Text (avg)}} & \textcolor{gray}{54.4}& \textcolor{gray}{51.4}& \textcolor{gray}{55.3} \\
Arc-Challenge & \textbf{36.3} & 31.6 & 35.6 \\
Arc-Easy & 64.0 & 60.2 & \textbf{65.3} \\
BoolQ & 55.8 & 59.8 & \textbf{61.3} \\
HellaSwag & 51.4 & 42.6 & \textbf{51.8} \\
Piqa & \textbf{72.5} & 69.3 & 72.0 \\
Siqa & \textbf{46.5} & 44.7 & 46.0 \\
Winogrande & 54.1 & 51.9 & \textbf{55.2} \\
\textcolor{gray}{\textit{Vision (avg)}} & \textcolor{gray}{37.6} & \textcolor{gray}{25.7} & \textcolor{gray}{37.0} \\
COCO-Captions & \textbf{62.9} & 43.4 & 62.6 \\
ChartQA & 13.3 & 8.4 & \textbf{13.6} \\
DocVQA & 48.1 & 33.0 & \textbf{50.6} \\
DocVQA-Hard & 30.8 & 18.5 & \textbf{32.6} \\
InfographicVQA & \textbf{24.4} & 19.4 & \textbf{24.4} \\
MMMU & \textbf{19.3} & 9.4 & 12.2 \\
TextVQA & \textbf{64.5} & 47.9 & 63.2 \\
\bottomrule
\end{tabular}
}
\caption{\textbf{Heterogeneous hardware} ML performance of a 2B dense model with 26B tokens using Decoupled DiLoCo ($M=8$) with a mix of TPUv6e and TPUv5p.}
\label{tab:tab_heterogeneous_hw}
\end{table}

We evaluate Decoupled DiLoCo on a heterogeneous cluster comprising four learners on TPUv5e and four on TPUv5p, adjusting the device count per learner to account for differences in HBM size. Due to these hardware and scaling differences, the slowest learners natively trailed the fastest by $18\%$. To rigorously stress-test the system, we further artificially injected an additional $10\%$ speed variance across all learners. By employing a minimal quorum size ($K=1$) alongside the adaptive grace window, we maximized compute utilization while matching the machine learning performance of a fully blocking, synchronous training setup ($K=M=8$).

\autoref{tab:tab_heterogeneous_hw} compares the downstream performance of this asynchronous, low-quorum approach ($K=1$) against the fully synchronous baseline ($K=8$). We demonstrate that the adaptive quorum and its grace window effectively mask the speed discrepancies between the two TPU architectures. Consequently, the system achieves the same ML performance as a synchronous baseline without being bottlenecked by the speed of its slowest chips.  Moreover, we display the XProf \citep{openxla_xprof} of the XLA operations of training on learners of heterogeneous speeds in \autoref{fig:xprof_heterogeneous_speed}. On the top row, notice that with a quorum of $K=M=8$ the decoupled algorithm becomes blocking, resulting in gap where learners are idle. On the middle row, with a minimal quorum of $K=1$, there is no gap, but more often than not, the syncer only collects a quorum of 1 learner at a time. This results in the syncer performing much more frequent synchronizations as each learner finishes its step (bearing similarity with \citet{liu2024asynchronous}'s setting). On the bottom row, we use the minimal quorum $K=1$ but with the adaptive grace window, which allows the syncer to fetch a significant number of learners on each synchronization round despite learners running at different speeds.

Overall, by seamlessly integrating heterogeneous hardware into a single training run, Decoupled DiLoCo empowers practitioners to scavenge older chip generations for extra compute and enables smoother transitions during new hardware rollouts.

\subsection{Scalability}\label{sec:exp_scaling}

\begin{figure*}[ht]
    \centering
    \begin{subfigure}{\textwidth}
        \centering
        \includegraphics[width=0.95\textwidth]{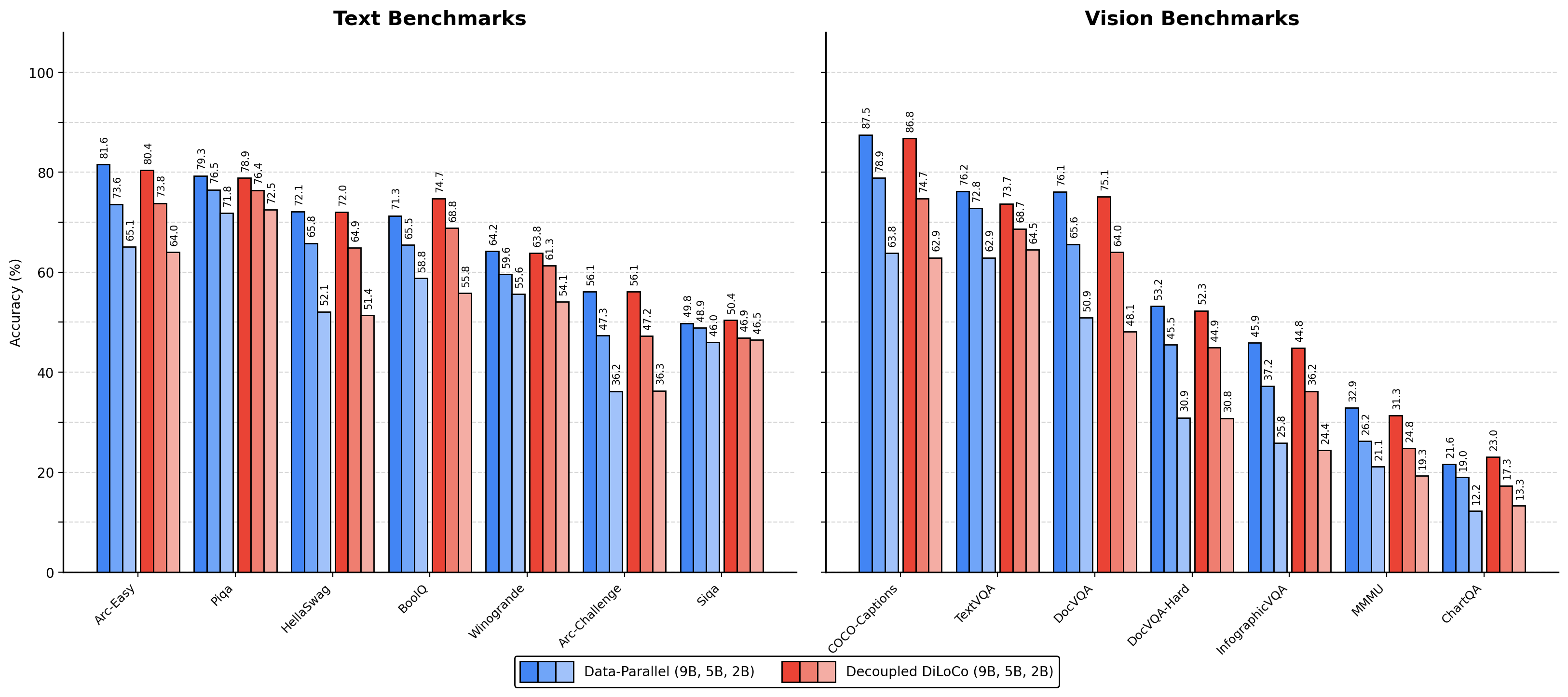}
        \caption{\textbf{Dense}: 2B, 5B, and 9B parameters models using 26B, 72B, and 141B tokens.}
        \label{fig:scaling}
    \end{subfigure}
    
    \vspace{1em} 
    
    \begin{subfigure}{\textwidth}
        \centering
        \includegraphics[width=0.95\textwidth]{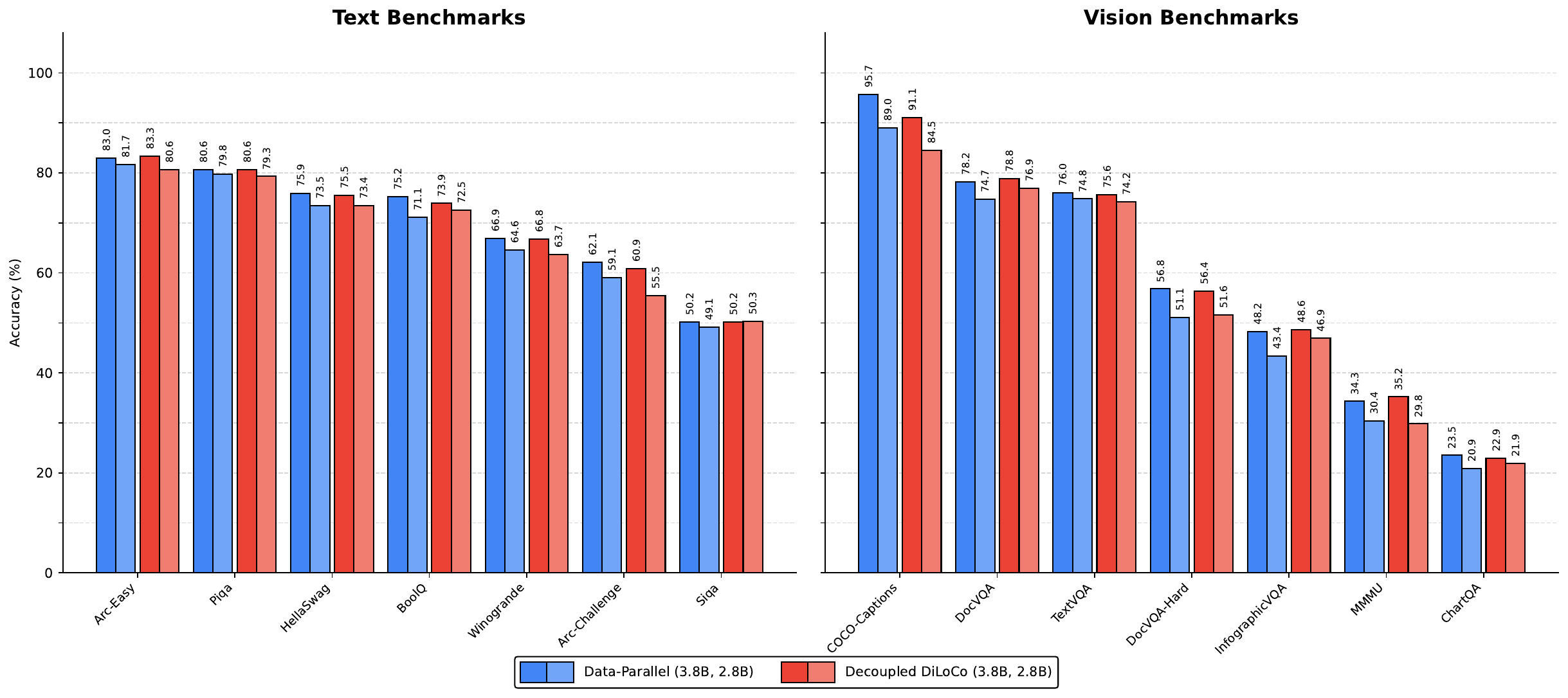}
        \caption{\textbf{Mixture-of-Experts}: 2.8B and 3.8B activated parameters models, using 170B and 233B tokens.}
        \label{fig:moe_scaling}
    \end{subfigure}
    
    \caption{\textbf{Downstream performances} for Dense (a) and Mixture-of-Experts (b) architectures.}
    \label{fig:combined_scaling}
\end{figure*}

Building on \citet{bergsma2025scalingcollapseefficientpredictable,qiu2025scalingcollapserevealsuniversal}, we tune to ensure predictable scaling of Decoupled DiLoCo across model scales.
Crucially, mirroring the observations of \citet{charles2025communication}, we observe that Decoupled DiLoCo exhibits predictable, improved performance as model size, batch size, and token budgets increase. Aligning with the ``bitter lesson'' \citep{sutton2019bitter}, that methods designed to exploit massive compute will prevail, our algorithm is ideally suited for large-scale pre-training environments where robust fault tolerance is a critical necessity.

\autoref{fig:scaling} reports the downstream performance of a standard data-parallel (DP) baseline versus Decoupled DiLoCo ($M=8$ learners) at scales of 2B, 5B, and 9B parameters dense models, trained for 26B, 72B, and 141B tokens, respectively. At every scale, Decoupled DiLoCo matches the performance of the centralized monolithic DP baseline. 

\autoref{fig:moe_scaling} extends this evaluation to a Mixture-of-Experts (MoE) architecture \citep{shazeer2017outrageously,fedus2022switch} with 2.8B and 3.8B active parameters trained on 170B and 233B tokens, respectively, once again demonstrating comparable downstream performance. We note that the success in this setting is potentially surprising - the learners use a load-balancing loss~\citep{shazeer2017outrageously}, but can only optimize this loss locally, not globally. Despite this, with sufficient scale we see that Decoupled DiLoCo with $M = 8$ performs comparably to SPMD data-parallel training.

\FloatBarrier 
\section{Conclusion}\label{sec:conclusions}

Decoupled DiLoCo allows us to decompose monolithic SPMD pre-training into independent asynchronous learners coordinated by a lightweight synchronizer.
By using minimum-quorum aggregation, token-weighted merging via Radial-Directional Averaging, balanced tensor fragmentation, and an adaptive grace window, the system prioritizes availability and partition tolerance without sacrificing downstream model quality. Our extensive experiments above show that Decoupled DiLoCo matches data-parallel performance on text and vision benchmarks across dense and MoE architectures at scales up to 9B parameters, while maintaining 88\% goodput under aggressive simulated failures (versus 58\% for elastic data-parallel) and perfect uptime. The framework further enables seamless scavenging of opportunistic compute and integration of heterogeneous hardware, unlocking the next frontier of cluster, distributed across geo-locations and chip generations.

We believe that Decoupled DiLoCo points at a new direction that upends the SPMD pre-training paradigm. If we relax model synchronization requirements, we can extract more useful computation from failure-prone hardware. This vision for non-SPMD pre-training mirrors that presented by Pathways~\citep{barham2022pathways}, and in fact uses the capabilities of Pathways to make it a reality.

Decoupled DiLoCo performs better with scale relative to data-parallel training in both model quality and goodput. This observation is crucial: the pre-training settings that most benefit from Decoupled DiLoCo for system reasons (bandwidth and hardware reliability, exacerbated by the number of chips involved) are exactly the settings in which model quality is most comparable to SPMD training.

As pre-training expands to geo-distributed clusters and environments where both bandwidth and hardware reliability are severely constrained~\citep{aguerayarcas2025future}, we believe availability-first training will shift from advantageous to necessary.

\clearpage

\begin{strip}
\section*{Acknowledgements}

This work was done by a team of members across Google DeepMind and Google Research.

\begin{multicols}{4}

\noindent \textbf{Leads}
\begin{itemize}[label={}, leftmargin=0pt, nosep]
    \item Arthur Douillard
    \item Keith Rush
    \item Yani Donchev
    \item Zachary Charles
\end{itemize}
\columnbreak 

\noindent \textbf{Core Contributors}
\begin{itemize}[label={}, leftmargin=0pt, nosep]
    \item Ayush Dubey
    \item Blake Woodworth
    \item Ionel Gog
    \item Jenny Bishop
    \item Josef Dean
    \item Nate Keating
    \item Nova Fallen
    \item Zachary Garrett
\end{itemize}
\columnbreak 

\noindent \textbf{Partial Contributors}
\begin{itemize}[label={}, leftmargin=0pt, nosep]
    \item Arthur Szlam
    \item Edouard Yvinec
    \item Henry Prior
\end{itemize}
\columnbreak 

\noindent \textbf{Advisors \& Sponsors}
\begin{itemize}[label={}, leftmargin=0pt, nosep]
    \item Brendan McMahan
    \item Chase Hensel
    \item Daniel Ramage
    \item Jeff Dean
    \item Lechao Xiao
    \item Marc'Aurelio Ranzato
    \item Michael Isard
    \item Paul Barham
    \item Raia Hadsell
    \item Zoltan Egyed
\end{itemize}

\end{multicols}
\vspace{0.5cm} 
\hrule 
\end{strip}

\bibliography{main}

\clearpage
\appendix

\section{Related Work}\label{sec:related_work}

\paragraph{Distributed training at scale.} The scale of LLM pre-training has necessitated advancements in distributed training methods. One such line of work focuses on more efficient ways to shard data-parallel training across large numbers of accelerators. This includes work on distributed data-parallel training~\citep{shazeer2018mesh, li2020pytorch}, ZeRO and fully-sharded data parallelism~\citep{rajbhandari2020zero,ren2021zero,FairScale2021,zhao2023pytorch}, pipeline parallelism~\citep{petrowski1993performance,huang2019gpipe,narayanan2019pipedream}, model parallelism~\citep{dean2012large}, tensor parallelism~\citep{narayanan2021efficient,korthikanti2023reducing}, sequence parallelism~\citep{korthikanti2023reducing}, and (for mixture-of-experts models) expert parallelism~\citep{lepikhingshard}. To adapt these synchronous methods to heterogeneous hardware, systems like Sailor \citep{strati2025sailorautomatingdistributedtraining} automate the search for load-balanced parallelization configurations.

\paragraph{DiLoCo and related methods.} An alternative approach to pre-training involves using periodic synchronization across accelerators, reducing network bottlenecks in training. This idea has existed for decades~\citep{mangasarian1993backpropagation}, and has been periodically re-purposed (or re-discovered) in a variety of settings, including federated learning~\citep{zinkevich2010parallelized, mcmahan2017communication, stich2018local}. The use of inner and outer optimizers was first proposed by \citet{hsu2019measuring} and \citet{reddi2021adaptive} in the context of federated learning, focusing on SGD as the inner optimizer and SGDM or Adam~\citep{kingma2014adam} as the outer optimizer. This type of approach was used by \citet{douillard2023diloco} for language model pre-training, where it was adapted specifically for this setting by using Adam as the inner optimizer and SGDM with Nesterov momentum in the outer. Since then, there has been a large amount of work building on this, including versions of DiLoCo with better system properties~\citep{liu2024asynchronous,kale2025eager,qi2025dilocox,kim2025halos}, variants of DiLoCo that further reduce bandwidth consumption~\citep{douillard2025streaming,beton2025improving,sarfi2025communication}, empirical scaling-oriented analyses of DiLoCo~\citep{charles2025communication,therien2025muloco}, and other algorithmic variants of DiLoCo~\citep{sani2024photon,therien2025muloco,kolehmainen2025noloco,sarfi2025communication,fan2025pier,lidin2026covenant72bpretraining72bllm}.

As discussed above, DiLoCo shares a conceptually similar underpinning to many optimization algorithms in federated learning, especially the FedOpt framework~\citep{reddi2021adaptive}. While we defer to~\citep{wang2021field} for an overview of federated optimization, we highlight a number of connections particularly relevant to Decoupled DiLoCo. First, we note that our token-based weighting strategy to account for variable speeds is similar to the step-weighting schema proposed by~\citet{wang2020tackling}. Second, we note that the idea of minimum quorum is analogous to over-provisioning client sampling in federated learning, something used in production settings by~\citet{bonawitz2019towards}. Last, we note that many of the asynchrony-aware techniques used by Decoupled DiLoCo are conceptually similar to the Papaya framework~\citep{huba2022papaya} for federated learning.

DiLoCo shares many similarities with work on communication-efficient training and optimization methods for LLMs. These include techniques like quantization~\citep{alistarh2017qsgd}, sparsification~\citep{wang2023cocktailsgd}, and communication-efficient optimizers~\citep{ahn2025dion,jovanovic2026lordo} though these approaches are typically orthogonal and can be combined with DiLoCo~\citep{douillard2025streaming}.

\paragraph{Asynchrony and fault-tolerance at LLM scales.} Parallel to DiLoCo, there are a number of works that that provide robustness to hardware failures and stragglers at LLM scales by employing forms of asynchrony and fault tolerance. There are largely two axes to parallelize over when employing such techniques: the layer/pipeline axis, or the data axis. For the former, \citet{borzunov2023distributed} and \citet{ryabinin2023swarm} train LLMs over unreliable networks by dynamically re-routing computations through the model in the event of stragglers and failures. Similarly, \citet{jang2023oobleck} provide resilience by reconfiguring pipeline templates in after node failures. Such approaches have also sparked research on designing optimizers for asynchronous pipeline parallelism~\citep{ajanthan2025momentum}.

On the data axis, much of the work (including our own) is similar in spirit to Hogwild!~\citep{niu2011hogwildlockfreeapproachparallelizing}. Building on the parameter server based formulation of asynchronous stochastic gradient descent in \citet{dean2012large}, \citet{zhang2016staleness} adapt learning rates based on staleness metrics. Unlike variants of asynchronous SGD, the Decoupled DiLoCo approach to asynchrony never requires updating the model using a stale gradient computed on a previous version of the parameters. Other works introduce fault-tolerant implementations of all-reduce.  \citet{ryabinin2021moshpit} utilize a butterfly all-reduce to handle unreliable devices, and more recently, \citet{salpekar2026training} use fault-tolerant sharded data-parallelism for resilience in large-scale distributed jobs. The use of asynchrony in tandem with DiLoCo was first proposed by~\citet{liu2024asynchronous}, and was extended to reduce inter-region communication by~\citet{kim2025halos}.

\section{Post-training}\label{sec:post_training}

To ensure that the benefits of Decoupled DiLoCo during pre-training do not come at the cost of downstream capabilities, we evaluate the impact of our framework on the post-training phase. We take the 5B parameter models trained on 1T tokens from our experiments with simulated hardware failures (see \autoref{sec:exp_hardware_failures}) and subject them to an identical post-training pipeline, which is a light version of Gemma 4 recipe.

Specifically, we compare three pre-training setups: a standard Data-Parallel (DP) baseline in an ideal environment with no failures, Decoupled DiLoCo in an ideal environment, and Decoupled DiLoCo trained under heavy simulated hardware failures ($\texttt{MTBI}_\text{chip} = 1$ year, $N_\text{chip}=1.2\text{m}$). We apply the same light version of the Gemma 4 post-training recipe to all three pre-trained checkpoints to observe how they adapt. 

The results, presented in Table~\ref{tab:post_training}, demonstrate that Decoupled DiLoCo maintains highly competitive post-training performance. We note that this finding is contrary to that of~\citet{acker2025happensnanochatmeetsdiloco}, where their iteration on DiLoCo was worse than data-parallel training even in the pre-training phase. We believe that this indicates the importance of making DiLoCo work as well as data-parallel in pre-training phases, and that post-training does not close any gaps in model quality at this phase.

Across evaluations spanning knowledge (MMLU-Pro \citep{wang2024mmluprorobustchallengingmultitask}), multi-lingual understanding (GMMLU-lite \citep{singh2024globalmmluunderstandingaddressing}), mathematical reasoning (GSM8K \citep{cobbe2021gsm8k}), and coding (HumanEval \citep{chen2021codex}), the models pre-trained with Decoupled DiLoCo perform comparably to, and in several cases, noticeably exceed, the monolithic DP baseline. Crucially, this parity holds even for the model pre-trained under chaotic, asynchronous hardware failures, confirming that the relaxed synchronization and failure recovery mechanisms of Decoupled DiLoCo do not compromise the model's capacity for post-training.

\begin{table*}[htbp]
\centering
\begin{tabular}{l c cccc}
\toprule
 & \textbf{Simulated} & \textbf{MMLU-Pro} & \textbf{GMMLU} & \textbf{GSM8K} & \textbf{HumanEval} \\
\textbf{Pre-training Method} & \textbf{HW Fails?} &
{\scriptsize\color{gray} knowledge} &
{\scriptsize\color{gray} multi-lingual} &
{\scriptsize\color{gray} math} &
{\scriptsize\color{gray} code} \\
\midrule
Data-Parallel     & No  & \textbf{21.4} & 42.3 & 23.8 & \underline{22.0} \\
Decoupled DiLoCo  & No  & \underline{18.6} & \textbf{43.3} & \textbf{35.4} & \textbf{22.6} \\
Decoupled DiLoCo  & Yes & 18.3 & \underline{42.4} & \underline{28.7} & \underline{22.0} \\
\bottomrule
\end{tabular}
\vspace{2mm}
\caption{\textbf{Post-training} results for Gemma 5B models, pretrained on 1T tokens. All models were pre-trained under their respective setups and subsequently fine-tuned using the same light Gemma 4 post-training recipe. Results indicate that Decoupled DiLoCo pre-training does not degrade post-training performance, even when subjected to hardware failures.}
\label{tab:post_training}
\end{table*}

\section{Fragmentation strategies}\label{sec:fragmentation_ablations}

Recall the fragmented nature of \autoref{alg:learner} and \autoref{alg:syncer}. Communication to and from the syncer operates on fragments of the model. Thus, the fragmentation strategy (how we partition the model weights into $P$ fragments) has two distinct potential forms of impact on the performance of these algorithms.
\begin{enumerate}
    \item The fragmentation strategy can impact the \textbf{model performance} (measured via various downstream tasks).
    \item The fragmentation strategy can impact the \textbf{system performance}, (by changing the bandwidth requirement of different communication steps).
\end{enumerate}

Perhaps surprisingly, we find that downstream model performance is quite robust to fragmentation strategy, and therefore we opt for fragmentation strategies that have maximally beneficial bandwidth usage profiles. To illustrate this, we describe the various fragmentation strategies we tried.

\paragraph{Layer fragmentation}

The first strategy is that proposed by \citet{douillard2025streaming}. In this, we split a model into fragments according to its layer structure. First, we have a single fragment for all non-transformer layers. Let $L$ denote the number of transformer layers. For $P - 1 \leq L$, we form the remaining $P - 1$ fragments by strided selection. For example, if $L = 6$ and $P -1 = 3$, then transformer layers 1 and 4 form a fragment, as do transformer layers 2 and 5, and 3 and 6.

\begin{figure*}[t]
    \captionsetup[subfigure]{justification=centering}
    \centering
    
    \begin{subfigure}{\linewidth}
        \centering
        \includegraphics[width=\linewidth]{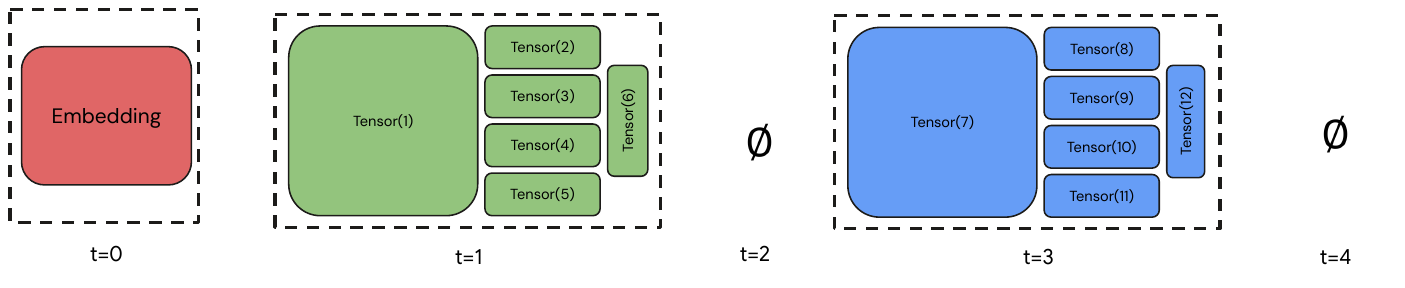}
        \caption{Layer Fragmentation. Because $H$ is larger than the number of layers, when $t = 2, 4$, we do not send any fragment to the syncer, resulting in bursty communication with potentially high peak bandwidth usage.}
        \label{fig:layer_fragments}
    \end{subfigure}

    \vspace{1em}

    \begin{subfigure}{\linewidth}
        \centering
        \includegraphics[width=\linewidth]{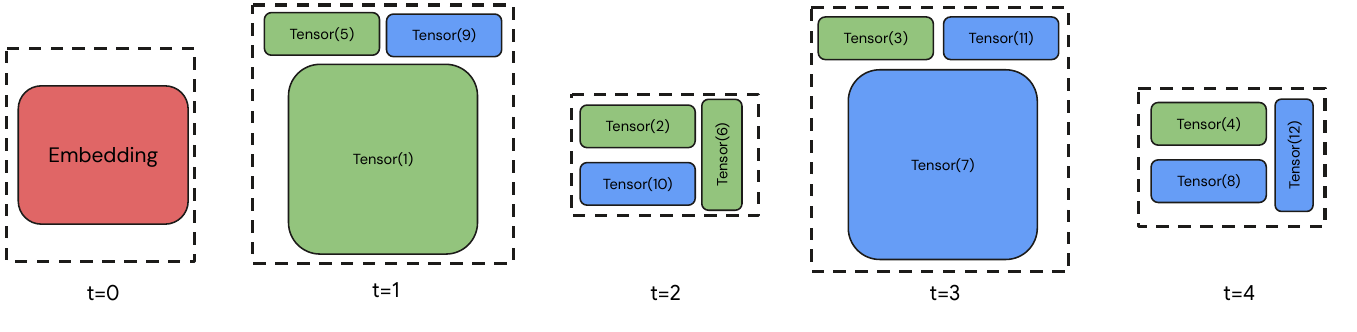}
        \caption{Tensor Fragmentation. We send a fragment to the syncer at every step, but the fragments have drastically different sizes, so the peak bandwidth usage might still be high.}
        \label{fig:tensor_fragments}
    \end{subfigure}

    \vspace{0.5em}

    \begin{subfigure}{\linewidth}
        \centering
        \includegraphics[width=\linewidth]{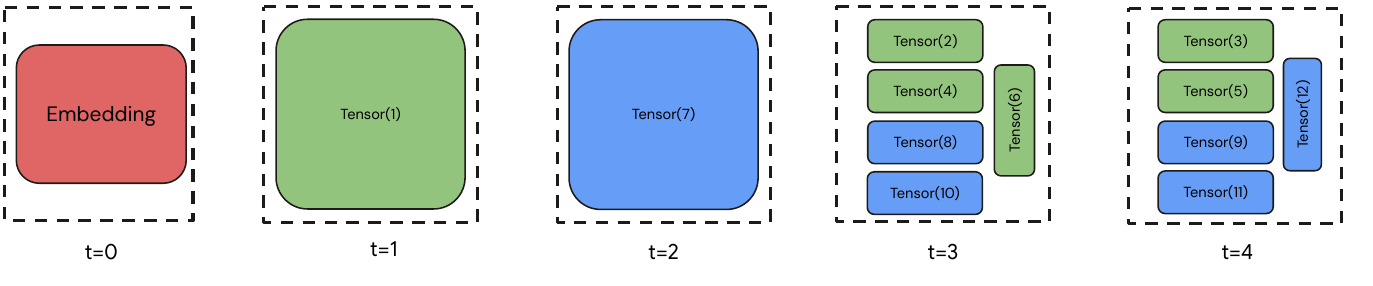}
        \caption{Balanced Tensor Fragmentation. This sends a fragment to the syncer at every step, with approximately equal sizes, reducing peak bandwidth and avoiding bursty communication.}
        \label{fig:balanced_tensor_fragments}
    \end{subfigure}
    \caption{Layer, tensor, and balanced tensor fragmentation applied to a model with an embedding layer and 2 transformer layers, each containing 6 tensors. We set $H=5$, and visualize which fragment is sent from the learner to the syncer at every step.}
    \label{fig:total_comparison}
\end{figure*}

While this strategy yields good training performance~\citep{douillard2025streaming}, layer fragmentation has system drawbacks. If the number of transformer layers $Q$ is less than $H$, then there will necessarily be some steps at which we do not communicate with the syncer. This means that the communication incurred can be quite bursty. See \autoref{fig:layer_fragments} for a simple example of this.

\paragraph{Tensor fragmentation}

To remedy this, we can instead partition model at the level of individual tensors (as opposed to transformer layers, which typically contain multiple tensors). In this strategy, we put all non-transformer tensors in the first fragment. The remaining tensors (say there are $S$ of them) are bucketed into $P - 1$ fragments in a strided manner by their index in some ordering of the tensors. For example, if $S = 9$ and $P - 1 = 3$, then (in addition to the fragment  with non-transformer tensors), the first fragment will have tensors $1, 4, 7$, the second will have tensors $2, 5, 8$, and the third will have tensors $3, 6, 9$. 

While this allows us to communicate smaller fragments more often (avoiding the extremely bursty communication patterns of layer fragmentation), the fragments may still be of very different sizes, meaning that bandwidth usage is still uneven. See \autoref{fig:tensor_fragments} for an example.

\paragraph{Balanced tensor fragmentation}

The last strategy we examine attempts to make the fragments as evenly balanced, in terms of total size, as possible. To do so we use a greedy bin-packing algorithm to pack the $Q$ tensors in the model into $P$ buckets. Specifically, we sort the tensors by their total size (measured in bits), and pack them into bins according via greedy number partitioning, where we iterate over the tensors in descending order of size, and put each successive tensor in the fragment whose total size is the smallest.

This has many beneficial system properties. First, as long as $Q \geq H$, we are always guaranteed to have $P = H$, so that we communicate a fragment at every step. Moreover, the greedy algorithm is well-known to be no more than $4/3$ times worse than the optimal packing strategy, measured in terms of the maximum size of any fragment~\citep{graham1969bounds}. This means that the peak bandwidth usage of balanced tensor fragmentation is at most $4/3$ of the minimum possible peak bandwidth of any fragmentation strategy. See \autoref{fig:balanced_tensor_fragments} for an example.

\paragraph{Sub-tensor fragmentation}

In preliminary investigations, we also evaluated a \emph{sub-tensor} fragmentation strategy, where we would split individual tensors across fragments. These were essentially split across the output of the tensor (e.g. the columns of the embedding layer), to guarantee that inputs were not effectively partitioned across fragments. By construction, this yields fragments of even more balanced size than the balanced tensor strategy above. While we found it competitive with other schemes, we opted not to use it in final experiments due to implementation complexity and to aid in debugging.

\subsection{Fragment Size Comparison}

\begin{figure*}[t]
    \captionsetup[subfigure]{justification=centering}
    \centering
    
    \begin{subfigure}{0.48\linewidth}
        \centering
        \includegraphics[width=\linewidth]{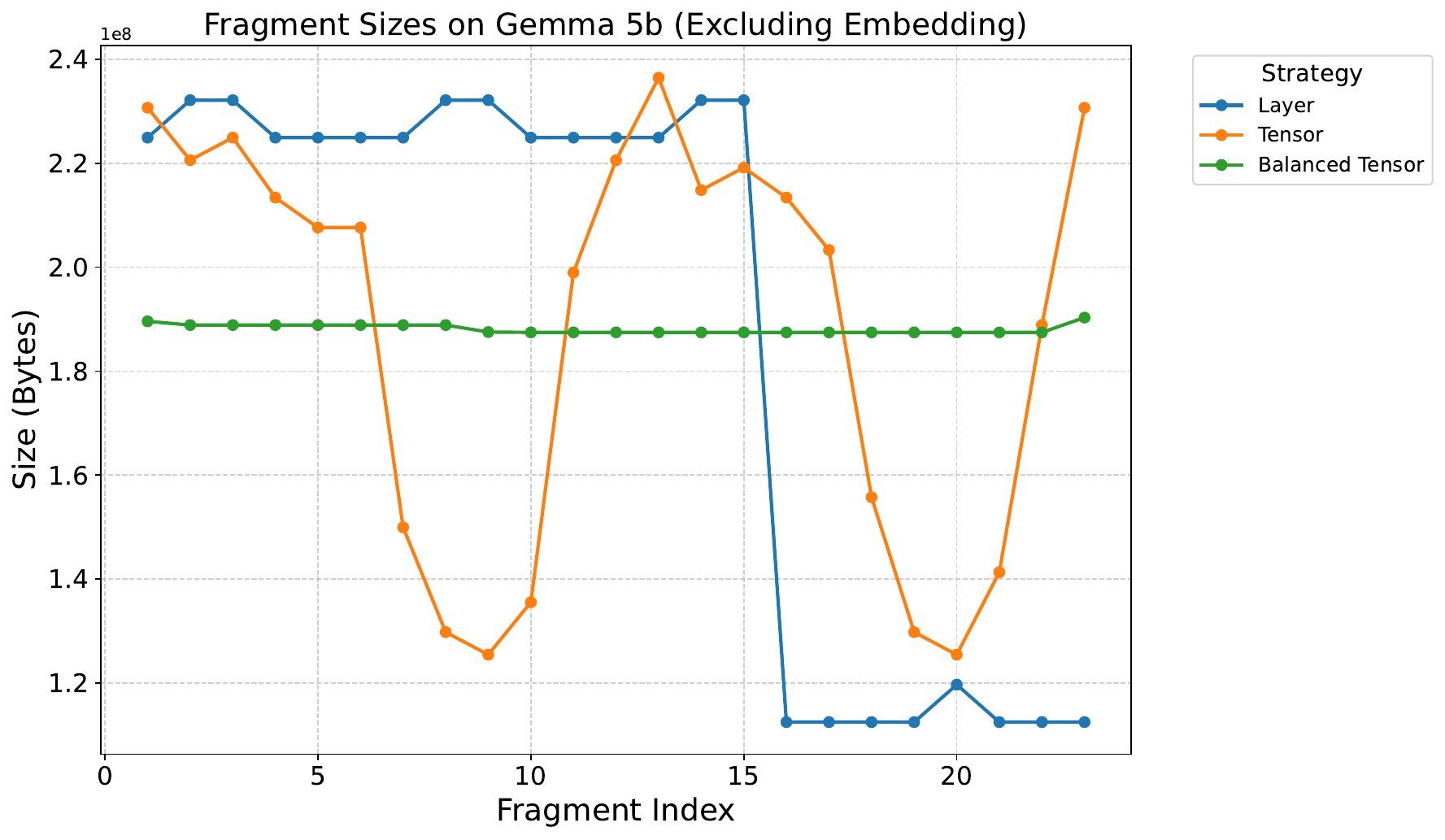}
        \caption{Gemma 5B.}
        \label{fig:gemma_5b_fragment_sizes}
    \end{subfigure}
    \hfill
    \begin{subfigure}{0.48\linewidth}
        \centering
        \includegraphics[width=\linewidth]{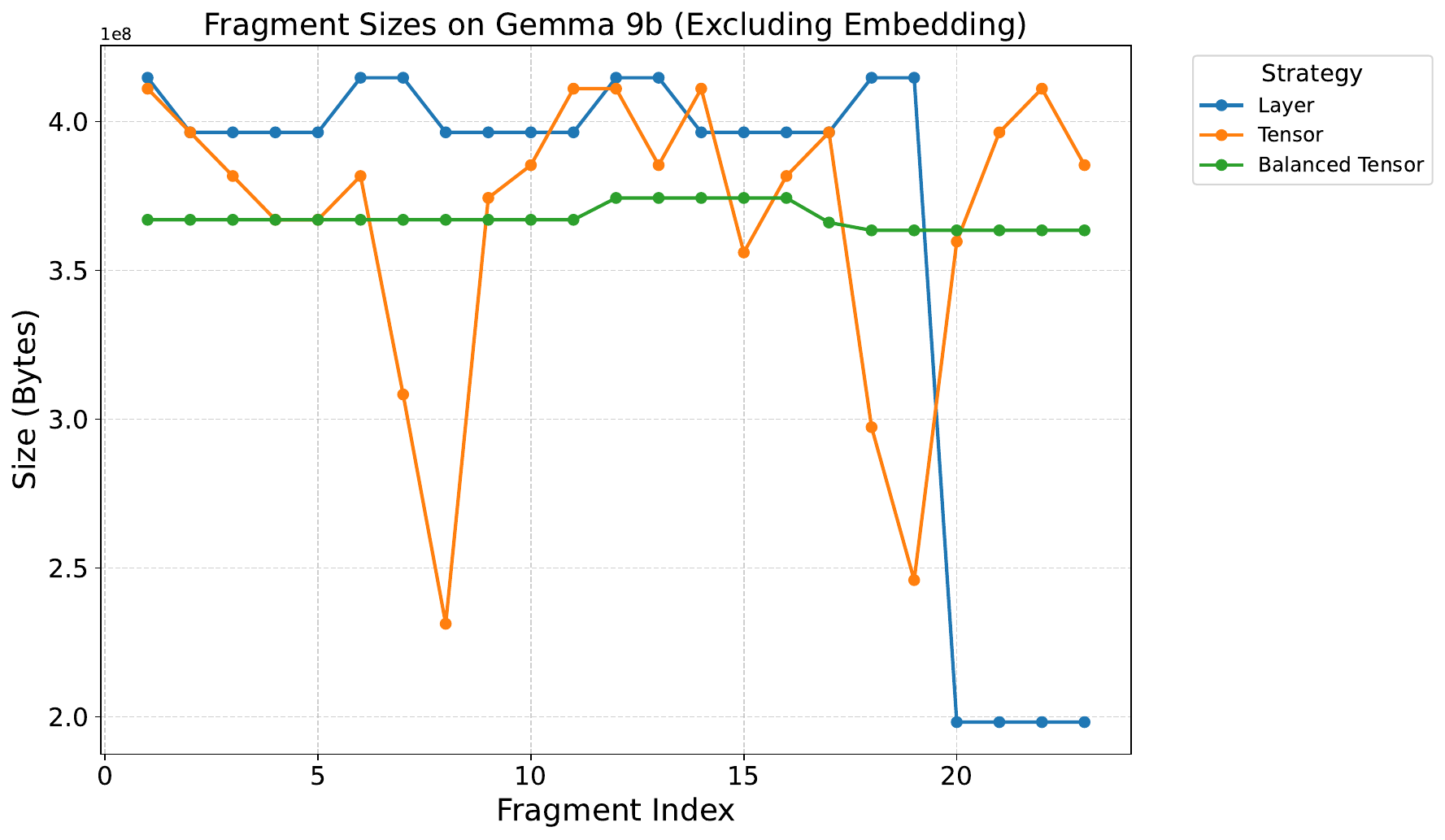}
        \caption{Gemma 9B.}
        \label{fig:gemma_9b_fragment_sizes}
    \end{subfigure}
    
    \caption{Comparison of fragment sizes on dense 5B and 9B models. Here we omit the embedding layer for visual clarity, as in all three cases it is in its own fragment that is significantly larger.}
    \label{fig:dense_fragment_sizes}
\end{figure*}

We now compare the size of fragments created by applying the above strategies to various models. Here we set $P = 24$ for all three methods, and record the total size (in bytes) of each fragment. This is simply the sum of the size of each tensor that is grouped under the same fragment. We show the results for the 5B and 9B dense models in \autoref{fig:dense_fragment_sizes}. 

\begin{figure*}[t]
    \captionsetup[subfigure]{justification=centering}
    \centering
    
    \begin{subfigure}{0.48\linewidth}
        \centering
        \includegraphics[width=\linewidth]{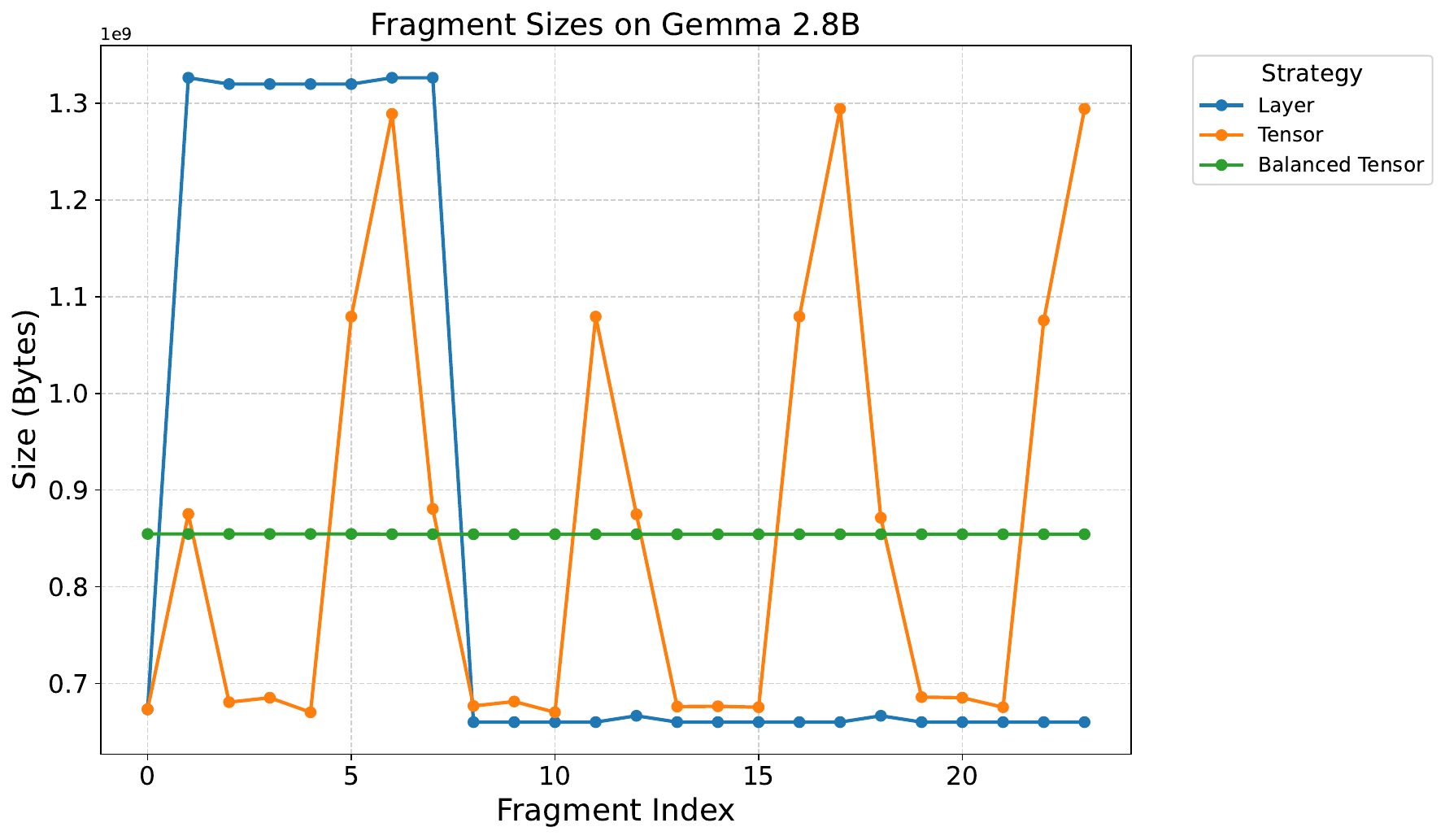}
        \caption{Gemma 2.8B MoE.}
        \label{fig:gemma_2p8b_fragment_sizes}
    \end{subfigure}
    \hfill
    \begin{subfigure}{0.48\linewidth}
        \centering
        \includegraphics[width=\linewidth]{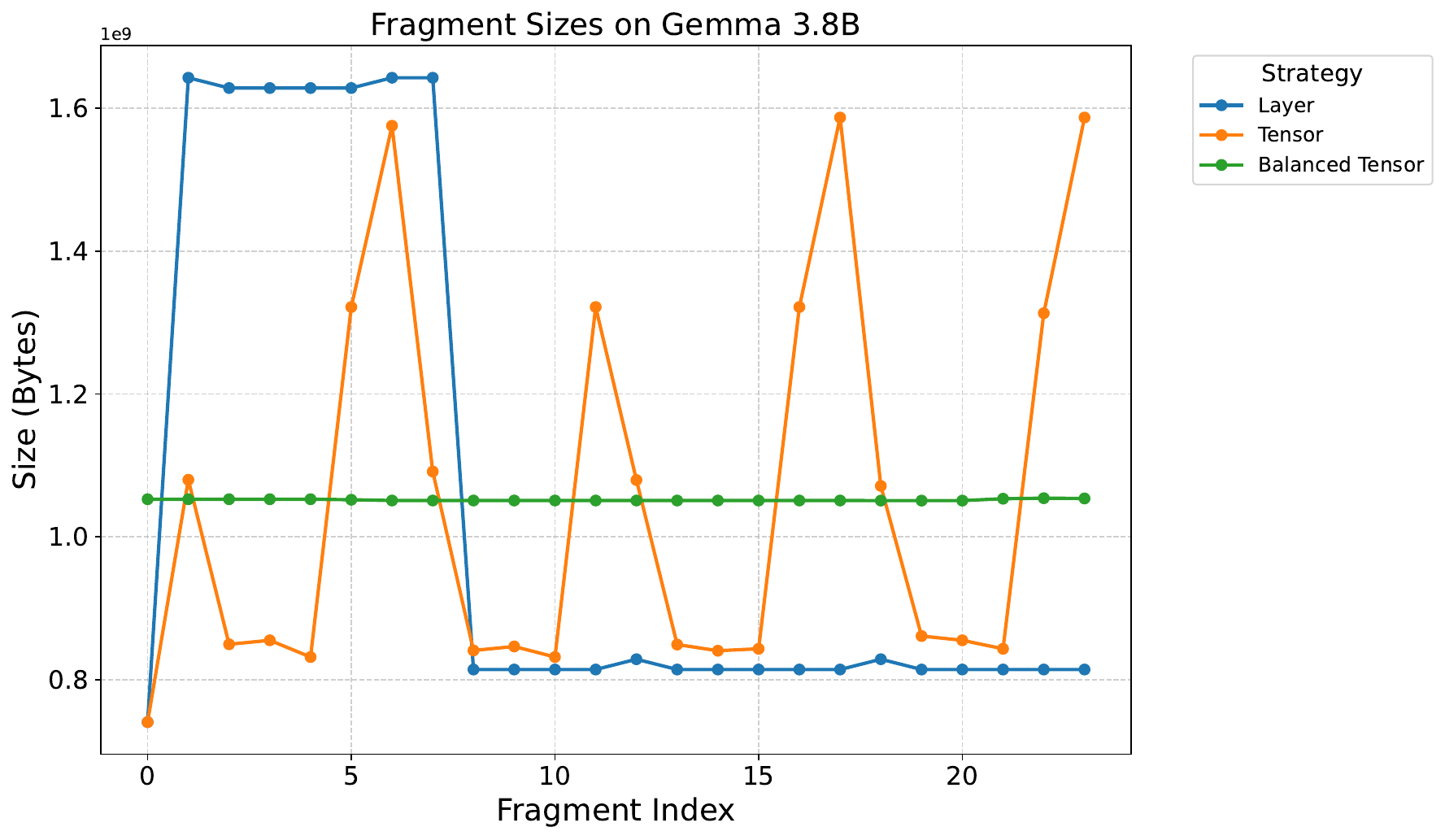}
        \caption{Gemma 3.8B MoE.}
        \label{fig:gemma_3p8b_fragment_sizes}
    \end{subfigure}
    
    \caption{Comparison of fragment sizes on Mixture-of-Experts models with 2.8B and 3.8B activated parameters.}
    \label{fig:moe_fragment_sizes}
\end{figure*}

\subsection{Evaluation Results}

We compare the three strategies on a 2B parameter model, setting $H = P = 24$, and $M = 8$. See \autoref{tab:fragmentation_ablation_transposed} for the results. We find that the methods all perform quite comparably on downstream evaluations. This is potentially surprising - tensor and balanced tensor fragmentation do not respect the layer structure of the model, but this does not reduce model quality. Thus, the only factor that is important then is their system performance. Because balanced tensor fragmentation has the lowest peak bandwidth and most consistent bandwidth usage across fragments, we use this strategy for all other experiments. 


\begin{table*}[t]
    \centering
    \small
    \setlength{\tabcolsep}{3.2pt}
    \resizebox{\textwidth}{!}{
        \begin{tabular}{l cccccccc cccccccc}
            \toprule
             & \multicolumn{8}{c}{Text Benchmarks} & \multicolumn{8}{c}{Vision Benchmarks} \\
            \cmidrule(lr){2-9} \cmidrule(lr){10-17}
            Method & \textcolor{gray}{\scriptsize Avg} & Arc-C & Arc-E & BoolQ & Hella & Piqa & Siqa & Wino & \textcolor{gray}{\scriptsize Avg} & COCO & Chart & Doc & Doc-H & Info & MMMU & TextVQA \\
            \midrule
            Balanced      & \textcolor{gray}{\scriptsize 54.4} & 36.3          & 64.0          & 55.8          & 51.4          & \textbf{72.5} & \textbf{46.5} & 54.1          & \textcolor{gray}{\scriptsize 37.6} & \textbf{62.9} & 13.3          & 48.1          & \textbf{30.8} & 24.4          & \textbf{19.3} & \textbf{64.5} \\
            Layer         & \textcolor{gray}{\scriptsize 55.3} & \textbf{36.5} & 64.6          & 59.1          & \textbf{52.0} & 71.5          & \textbf{46.5} & \textbf{56.7} & \textcolor{gray}{\scriptsize 35.5} & 59.5          & 12.4          & 46.7          & 28.7          & \textbf{24.9} & 14.9          & 61.1          \\
            Tensor        & \textcolor{gray}{\scriptsize 55.4} & 35.8          & \textbf{65.8} & \textbf{61.6} & 51.8          & 72.0          & \textbf{46.5} & 54.5          & \textcolor{gray}{\scriptsize 35.8} & 57.1          & \textbf{15.0} & \textbf{48.8} & 30.6          & 24.1          & 13.9          & 61.2          \\
            \bottomrule
        \end{tabular}
    }
    \caption{Downstream evaluations of Gemma 2B trained on 26B tokens with Decoupled DiLoCo using M = 8 learners and various fragmentation strategies.}
    \label{tab:fragmentation_ablation_transposed}
\end{table*}

\section{Algorithmic Ablations}

Below we briefly detail various design choices we made when concretizing \autoref{alg:learner} and \autoref{alg:syncer}.

\subsection{Mixing learner and syncer params}\label{sec:interpolation_ablations}

In Algorithm \ref{alg:learner}, upon receiving $\Theta_p^{(t-\tau)}$ from the syncer, a learner simply sets their local fragment $\theta_p^{(t)}$ to this value. We could instead interpolate via
\begin{equation}  
\theta_p^{(t)} \gets \alpha \theta_p^{(t)} + (1-\alpha) \Theta_p^{(t)}\,.
\end{equation}
While \citet{douillard2025streaming} showed that $\alpha=0.5$ was useful for $M=2$ and $\tau \ge 10$, we found that at larger $M$ $\alpha=0$ is optimal. Intuitively, with more learners $M$, each learner contribution $\theta_{m, p}^{(t)}$ is less informative, and thus the global update $\Theta_p^{(t)}$ yields better performance.

\subsection{Merging methods}\label{sec:merging_ablations}

In Algorithm \ref{alg:syncer}, the syncer is responsible for merging the per-fragment learner updates. Here we discuss two methods for doing this merging, and how they compare.

\paragraph{Direct averaging} One option for the \mergefn function in \autoref{alg:syncer} is to mirror \citep{douillard2024diloco} and directly average the per-fragment outer gradients $\Delta_{m, p}^{(t)}$ from \eqref{eq:streaming_outer_opt}. Ignoring weighting, we would therefore define
\begin{align*}
& \mergefn\left(\{\theta_{m, p}^{(t)}\}_{i \in \mathcal{M}_t}, \Theta_{p}^{(t-H)}\right) \\
& = \dfrac{1}{|\mathcal{M}_t|} \sum_{i\in\mathcal{M}_t} \Delta_{m, p}^{(t)} = \dfrac{1}{|\mathcal{M}_t|} \sum_{i\in\mathcal{M}_t} \left(\Theta_p^{(t-H)} - \theta_{m, p}^{(t)} \right) \\
\end{align*}
The weighted version is defined analogously via
\begin{align*}
& \mergefn\left(\{\theta_{m, p}^{(t)}, w_{m, p}\}_{i \in \mathcal{M}_t}, \Theta_{p}^{(t-H)}\right) \\
& = \dfrac{\sum_{i\in\mathcal{M}_t} w_{m, p}\left(\Theta_p^{(t-H)} - \theta_{m, p}^{(t)} \right)}{\sum_{i \in \mathcal{M}_t} w_{m, p}} \\
\end{align*}

\paragraph{Radial-directional averaging.} Empirically, the per-learner outer gradients $\Delta_{m, p}^{(t)}$ are almost pairwise orthogonal. Thus, if they are all of norm $R$, their average has norm of approximately $R/\sqrt{M}$. Hence, we generally need to re-tune the outer optimizer when we merge via averaging.

To mitigate this, we can instead attempt to merge in a way that the output norm is essentially invariant to $M$. We propose our own lightweight merging method, Radial-Directional Averaging (RDA), which separately averages the norm and direction of the vectors. Formally, RDA operates as follows. Define
\[
\phi: \mathbb{R}^n\backslash\{0\} \to \mathbb{R}^n, x \mapsto \dfrac{x}{\|x\|_2}.
\]

For a set of non-zero vectors $v_1, \dots, v_M$ we define the function $\rda$ via:
\[
\rda(v_1, \dots, v_M) = \left(\dfrac{1}{M}\sum_{i=1}^M \|v_i\|_2 \right) \phi\left(\dfrac{1}{M}\sum_{i=1}^M \phi(v_i) \right).
\]
That is, we average the norm and direction of the vectors separately, projecting the average direction to be a unit vector, and multiply them. Note that if $\|v_i\| = R$ for all $i$, then $\|\rda(v_1, \dots, v_M)\| = R$. We define a weighted version of $\rda$ completely analogously. Let $w_1, \dots, w_M$ denote the weights for each vector. Then
\begin{align*}
& \rda\left(\{(v_i, w_i)\}_{i=1}^M\right) = \left(\dfrac{\sum_{i=1}^M w_iv_i}{\sum_{i=1}^M w_i}\right) \phi\left(\dfrac{\sum_{i=1}^M w_i \phi(v_i)}{\sum_{i=1}^M w_i} \right)
\end{align*}

To specify the corresponding \mergefn function in \ref{alg:syncer}, we apply $\rda$ to the (weighted) per-learner outer gradients, as when we do direct averaging.

Note that at first glance, RDA requires two all-reduces while direct averaging only one. However, the magnitude tensors are less than $0.05\%$ of the size of the direction tensors, making the overhead negligible.

\paragraph{Other options}

For $M = 2$ one can use SLERP~\citep{shoemake1985animating}, but it is unclear how best to generalize this to $M > 2$. While~\citep{buss2001spherical} develop a generalization for $M > 2$, it involves least-squares minimization, and as such incurs more computational overhead than methods like direct averaging. \citet{rame2024warpbenefitsweightaveraged} propose an iterative version of SLERP to merge $M > 2$ tensors, but we found that RDA performed comparably if not better, with reduced complexity. We also evaluated other merging schemes, such as TIES-MERGING~\citep{yadav2023ties}, Iso-C~\citep{marczak2025taskleftbehindisotropic}, and sparsity-aware merging schemes, but found that direct averaging and RDA outperformed such methods in our setting. This is potentially due to the fact that much of the work on such merging methods has focused on merging after large numbers of steps~\citep{wortsman2022modelsoupsaveragingweights} on non-IID distributions, while we do partial model merging essentially every step.

\paragraph{Results}

We compare direct averaging (Avg) and RDA on a 2B parameter model. Qualitatively, we find that the outer gradients corresponding to the embedding component of the model often did not have the ``near-orthogonality'' property described above. Therefore, we compared the use of Avg and RDA separately on the embedding component and on the rest of the model. We compare this for $M = 8$ learners (\autoref{tab:merging_ablation_m_8}) and $M = 16$ learners (\autoref{tab:merging_ablation_m_16}).

For $M = 8$, there is no entirely consistent method that outperforms the others. However, we find that on nearly all tasks, applying RDA to the model and Avg to the embedding performs comparably or better to the other methods. For $M=16$, we find that applying RDA to the non-embedding component of the model drastically improves performance. Moreover, applying Avg to the embedding generally (though not uniformly) boosts performance slightly over RDA. Therefore, for all of our primary experiments we apply Avg to the embedding component, and RDA to the remaining parts of the model.

\begin{table*}[t]
    \centering
    \footnotesize
    \setlength{\tabcolsep}{2.5pt}
    \begin{tabular}{ll cccccccc cccccccc}
        \toprule
        \multicolumn{2}{c}{Merge Method} & \multicolumn{8}{c}{Text Benchmarks} & \multicolumn{8}{c}{Vision Benchmarks} \\
        \cmidrule(lr){1-2} \cmidrule(lr){3-10} \cmidrule(lr){11-18}
        Model & Embed & \textcolor{gray}{\scriptsize Avg} & Arc-C & Arc-E & BoolQ & Hella & Piqa & Siqa & Wino & \textcolor{gray}{\scriptsize Avg} & COCO & Chart & Doc & Doc-H & Info & MMMU & TextVQA \\
        \midrule
        Avg           & Avg           & \textcolor{gray}{\scriptsize 54.1} & 35.8          & 63.1          & 58.0          & 50.5          & 71.1          & 46.3          & 54.2          & \textcolor{gray}{\scriptsize 34.2} & 58.8          & 11.9          & 46.1          & 27.2          & 22.9          & 13.8          & 58.7          \\
        Avg           & RDA           & \textcolor{gray}{\scriptsize 53.3} & 34.6          & 59.7          & 59.8          & 48.3          & 71.2          & 45.2          & 54.5          & \textcolor{gray}{\scriptsize 32.9} & 53.7          & 12.1          & 40.7          & 24.8          & 20.7          & 19.2          & 59.0          \\
        RDA           & Avg           & \textcolor{gray}{\scriptsize 54.4} & \textbf{36.3} & \textbf{64.0} & 55.8          & \textbf{51.4} & \textbf{72.5} & \textbf{46.5} & 54.1          & \textcolor{gray}{\scriptsize 37.6} & \textbf{62.9} & \textbf{13.3} & \textbf{48.1} & \textbf{30.8} & 24.4          & \textbf{19.3} & \textbf{64.5} \\
        RDA           & RDA           & \textcolor{gray}{\scriptsize 55.1} & 34.7          & 63.6          & \textbf{62.8} & 50.9          & 72.3          & 46.2          & \textbf{55.1} & \textcolor{gray}{\scriptsize 35.6} & 59.4          & 12.6          & 46.0          & 28.4          & \textbf{24.6} & 16.1          & 62.2          \\
        \bottomrule
    \end{tabular}
    \caption{Downstream evaluations of Gemma 2B trained on 26B tokens with Decoupled DiLoCo using $M = 8$ learners and various merging methods.}
    \label{tab:merging_ablation_m_8}
\end{table*}

\begin{table*}[t]
    \centering
    \footnotesize
    \setlength{\tabcolsep}{2.5pt}
    \begin{tabular}{ll cccccccc cccccccc}
        \toprule
        \multicolumn{2}{c}{Merge Method} & \multicolumn{8}{c}{Text Benchmarks} & \multicolumn{8}{c}{Vision Benchmarks} \\
        \cmidrule(lr){1-2} \cmidrule(lr){3-10} \cmidrule(lr){11-18}
        Model & Embed & \textcolor{gray}{\scriptsize Avg} & Arc-C & Arc-E & BoolQ & Hella & Piqa & Siqa & Wino & \textcolor{gray}{\scriptsize Avg} & COCO & Chart & Doc & Doc-H & Info & MMMU & TextVQA \\
        \midrule
        Avg           & Avg           & \textcolor{gray}{\scriptsize 52.7} & 35.1          & 60.9          & 56.5          & 47.7          & 70.5          & 45.4          & 53.0          & \textcolor{gray}{\scriptsize 28.2} & 54.1          & 10.3          & 36.6          & 21.0          & 19.6          & 5.4           & 50.7          \\
        Avg           & RDA           & \textcolor{gray}{\scriptsize 49.3} & 29.7          & 52.3          & 56.6          & 40.6          & 68.7          & 45.3          & 51.8          & \textcolor{gray}{\scriptsize 22.4} & 31.6          & 9.7           & 26.4          & 15.6          & 15.2          & 14.0          & 44.4          \\
        RDA           & Avg           & \textcolor{gray}{\scriptsize 54.5} & 35.0          & \textbf{63.1} & \textbf{59.5} & \textbf{50.2} & \textbf{72.0} & 45.7          & \textbf{55.6} & \textcolor{gray}{\scriptsize 33.6} & \textbf{58.4} & \textbf{13.4} & 43.2          & \textbf{26.5} & 21.3          & 11.8          & \textbf{60.9} \\
        RDA           & RDA           & \textcolor{gray}{\scriptsize 52.7} & \textbf{35.6} & 60.9          & 53.3          & 48.6          & 71.4          & \textbf{46.3} & 52.8          & \textcolor{gray}{\scriptsize 33.9} & 56.6          & 12.2          & \textbf{43.3} & 26.2          & \textbf{24.2} & \textbf{15.3} & 59.2          \\
        \bottomrule
    \end{tabular}
    \caption{Downstream evaluations of Gemma 2B trained on 26B tokens with Decoupled DiLoCo using $M = 16$ learners and various merging methods.}
    \label{tab:merging_ablation_m_16}
\end{table*}

\subsection{Outer gradients compression}

Similarly to Streaming DiLoCo \citep{douillard2025streaming}, we can compress outer gradients from bf16 (16 bits) to int4 (4 bits) to minimize bandwidth constraint when doing the all-reduce synchronization. We display in \autoref{tab:grad_compresssion} the performance of a 2B dense model trained with high level of simulated hardware failures ($\texttt{MTBI}_\text{chip}=1$ year, and using $N_\text{chip}=1.2\text{m}$ simulated chips).  We found that compressing the outer gradients before their all-reduce (in the $\mergefn$, \texttt{L11} of \autoref{alg:syncer}) to int4 yields comparable performance to bf16, even in the asynchronous case with heavy rate of hardware failures. We also noticed that using 2 bits or 1 bit resulted in unacceptable performance regression. More exploration could be done to exploit more advanced compression scheme such as MuLoCo \citep{therien2025muloco}.

\begin{table}[htbp]
\centering
\resizebox{\columnwidth}{!}{
\begin{tabular}{lcccc}
\hline
\# bits / value  & 16 bits & 4 bits & 2 bits & 1 bit \\
\hline
Relative val loss & --- & +0.09\% & +32\% & +83\% \\
\midrule
\textit{\textcolor{gray}{Text (avg)}}        & \textcolor{gray}{55.4} & \textcolor{gray}{55.3} & \textcolor{gray}{42.6} & \textcolor{gray}{37.3} \\
Arc-C & 36.5 & 35.7 & 22.9 & 20.6 \\
Arc-Easy & 64.2 & 64.9 & 35.9 & 32.5 \\
BoolQ & 62.0 & 62.2 & 61.8 & 39.0 \\
HellaSwag & 51.8 & 51.5 & 27.2 & 25.4 \\
Piqa & 71.6 & 71.4 & 57.8 & 56.4 \\
Siqa & 45.8 & 46.6 & 40.7 & 37.1 \\
Winogrande & 55.6 & 54.8 & 51.7 & 50.4 \\
\midrule
\textcolor{gray}{\textit{Vision (avg)}}      & \textcolor{gray}{36.6}  & \textcolor{gray}{34.4} & \textcolor{gray}{5.3} & \textcolor{gray}{0.0} \\
COCO-Captions & 61.5 & 60.4 & 3.5 & 0.0 \\
ChartQA & 12.7 & 11.8 & 2.8 & 0.1 \\
DocVQA & 47.7 & 45.3 & 2.2 & 0.0 \\
DocVQA-Hard & 29.9 & 27.9 & 0.9 & 0.0 \\
InfographicVQA & 24.4 & 24.3 & 4.6 & 0.0 \\
MMMU & 18.0 & 9.9 & 17.6 & 0.1 \\
TextVQA & 62.0 & 61.3 & 5.8 & 0.0 \\
\hline
\end{tabular}
}
\caption{\textbf{Outer gradients compression} impact on ML performance for a 2B model trained on 26B tokens, while injecting hardware failures at the $\texttt{MTBI}_\text{chip} = 1$ year and $N_\text{chip}=1.2\text{m}$ simulated chips.}
\label{tab:grad_compresssion}
\end{table}

\subsection{Number of learners impact on asynchronous resiliency}

In \autoref{tab:tab_hw_failures_across_m}, we compare the ML performance on text and vision benchmark without simulated hardware failures and with ($N_\text{chip}=1.2\text{m}$ to $N_\text{chip}=2.4\text{m}$ simulated chips and $\texttt{MTBI}_\text{chip}$ = 1y, as in \autoref{sec:exp_hardware_failures}. Note that lower values of number of learners $M$ result in lower goodput because the blast radius is larger: for $M=2$, the radius is half of the total chips. Larger numbers of learners significantly improves goodput, here up to $M=16$ with 93\% goodput at $N_\text{chip}=1.2\text{m}$ chips. We show neutral ML performance from $M=2$ to $M=8$, and slightly degraded performance at $M=16$ but that can be mitigated by multiple aspects, including training larger models with larger token budget.

\begin{table*}[ht]
\centering
\setlength{\tabcolsep}{4pt} 
\begin{tabular}{l cccc cccc cccc}
\toprule
\multirow{2}{*}{} & \multicolumn{4}{c}{No HW failures} & \multicolumn{4}{c}{\begin{tabular}{@{}c@{}}With $N_\text{chip}=1.2\text{m}$ and \\ $\texttt{MTBI}_\text{chip}$ = 1y\end{tabular}} & \multicolumn{4}{c}{\begin{tabular}{@{}c@{}}With $N_\text{chip}=2.4\text{m}$ and \\ $\texttt{MTBI}_\text{chip}$ = 1y\end{tabular}} \\
\cmidrule(lr){2-5} \cmidrule(l){6-9} \cmidrule(l){10-13}
\# learners $M$ & 2 & 4 & 8 & 16 & 2 & 4 & 8 & 16 & 2 & 4 & 8 & 16 \\
\midrule
Goodput & 100\% & 100\% & 100\% & 100\% & 70\% & 82\% & 88\% & 93\% & 54\% & 73\% & 80\% & 86\% \\
\midrule
\textcolor{gray}{\textit{Text (avg)}} & \textcolor{gray}{55.1}& \textcolor{gray}{55.9}& \textcolor{gray}{54.4}& \textcolor{gray}{54.5}& \textcolor{gray}{54.4}& \textcolor{gray}{55.0}& \textcolor{gray}{55.4}& \textcolor{gray}{53.9}& \textcolor{gray}{54.4}& \textcolor{gray}{54.5}& \textcolor{gray}{54.6}& \textcolor{gray}{53.4} \\
Arc-Challenge & 37.1 & 36.0 & 36.3 & 35.0 & 35.7 & 36.9 & 36.5 & 34.4 & 35.4 & 35.5 & 35.1 & 35.8 \\
Arc-Easy & 64.1 & 65.2 & 64.0 & 63.1 & 64.2 & 63.5 & 64.2 & 63.5 & 61.1 & 63.0 & 64.4 & 62.6 \\
BoolQ & 58.2 & 61.2 & 55.8 & 59.5 & 59.4 & 60.2 & 62.0 & 59.8 & 59.2 & 60.2 & 60.2 & 54.1 \\
HellaSwag & 52.7 & 52.9 & 51.4 & 50.2 & 51.4 & 51.9 & 51.8 & 50.0 & 51.6 & 51.5 & 50.9 & 49.7 \\
Piqa & 72.4 & 72.2 & 72.5 & 72.0 & 71.9 & 71.5 & 71.6 & 71.0 & 71.8 & 71.7 & 70.9 & 71.2 \\
Siqa & 46.3 & 47.2 & 46.5 & 45.7 & 45.4 & 46.0 & 45.8 & 46.1 & 45.9 & 45.5 & 45.4 & 44.6 \\
Winogrande & 54.8 & 56.2 & 54.1 & 55.6 & 53.0 & 55.1 & 55.6 & 52.6 & 55.8 & 54.3 & 55.1 & 56.0 \\
\midrule
\textcolor{gray}{\textit{Vision (avg)}} & \textcolor{gray}{36.8} & \textcolor{gray}{36.8} & \textcolor{gray}{37.6} & \textcolor{gray}{33.6} & \textcolor{gray}{35.8} & \textcolor{gray}{35.6} & \textcolor{gray}{36.6} & \textcolor{gray}{35.1} & \textcolor{gray}{37.2} & \textcolor{gray}{35.3} & \textcolor{gray}{37.3} & \textcolor{gray}{34.3} \\
COCO-Captions & 62.8 & 58.6 & 62.9 & 58.4 & 59.5 & 61.8 & 61.5 & 56.8 & 59.0 & 59.8 & 63.1 & 55.0 \\
ChartQA & 12.0 & 14.0 & 13.3 & 13.4 & 14.1 & 13.4 & 12.7 & 13.6 & 13.5 & 12.4 & 13.5 & 12.5 \\
DocVQA & 47.9 & 50.8 & 48.1 & 43.2 & 48.1 & 47.9 & 47.7 & 47.1 & 47.8 & 48.3 & 50.4 & 46.8 \\
DocVQA-Hard & 28.6 & 31.1 & 30.8 & 26.5 & 28.5 & 28.7 & 29.9 & 26.8 & 30.8 & 25.6 & 30.7 & 28.2 \\
InfographicVQA & 26.4 & 24.9 & 24.4 & 21.3 & 23.8 & 24.0 & 24.4 & 23.2 & 24.2 & 23.8 & 25.1 & 23.0 \\
MMMU & 17.6 & 14.8 & 19.3 & 11.8 & 14.0 & 13.8 & 18.0 & 19.0 & 22.0 & 16.2 & 18.8 & 17.4 \\
TextVQA & 62.6 & 63.6 & 64.5 & 60.9 & 62.6 & 60.0 & 62.0 & 59.2 & 63.0 & 61.0 & 59.9 & 57.0 \\
\bottomrule
\end{tabular}
\caption{\textbf{Simulated hardware failures impact} for a 2B model trained on 26B tokens across different numbers of learners $M$, from 2 to 16 for 1.2 and 2.4 million simulated chips.}
\label{tab:tab_hw_failures_across_m}
\end{table*}

\subsection{Token budget impact}

We showcase in \autoref{tab:overtrain} the results of a 2B dense model trained with data-parallel and Decoupled DiLoCo methods with increasing token budget, at 26B tokens, 260B tokens ($10\times$), and 1.3T tokens ($50\times$). Notice that while Decoupled DiLoCo can be slightly weaker at low token budget, its performance surpasses data-parallel as the budget increases.

\begin{table*}[ht]
\centering
\setlength{\tabcolsep}{4pt} 
\begin{tabular}{l CCC CCC}
\toprule
\multirow{2}{*}{} & \multicolumn{3}{c}{Data-Parallel} & \multicolumn{3}{c}{Decoupled DiLoCo $M=8$} \\
\cmidrule(lr){2-4} \cmidrule(l){5-7}
\# tokens & 26B & 260B & 1.3T & 26B & 260B & 1.3T \\
\midrule
\textcolor{gray}{\textit{Text (avg)}} & \textcolor{gray}{55.1}& \textcolor{gray}{61.8}& \textcolor{gray}{64.5}& \textcolor{gray}{54.4}& \textcolor{gray}{62.1}& \textcolor{gray}{64.9} \\
Arc-Challenge & 36.2 & 46.2 & 51.1 & 36.3 & 48.1 & 51.5 \\
Arc-Easy & 65.1 & 73.2 & 75.5 & 64.0 & 73.8 & 77.0 \\
BoolQ & 58.8 & 66.1 & 72.2 & 55.8 & 66.1 & 71.6 \\
HellaSwag & 52.1 & 63.1 & 66.8 & 51.4 & 62.7 & 66.8 \\
Piqa & 71.8 & 76.6 & 77.3 & 72.5 & 76.0 & 77.8 \\
Siqa & 46.0 & 48.1 & 47.6 & 46.5 & 48.3 & 48.7 \\
Winogrande & 55.6 & 59.0 & 61.4 & 54.1 & 59.6 & 60.6 \\
\midrule
\textcolor{gray}{\textit{Vision (avg)}} & \textcolor{gray}{38.2} & \textcolor{gray}{47.8} & \textcolor{gray}{50.6} & \textcolor{gray}{37.6} & \textcolor{gray}{47.4} & \textcolor{gray}{52.1} \\
COCO-Captions & 63.8 & 75.2 & 78.5 & 62.9 & 74.2 & 83.7 \\
ChartQA & 12.2 & 17.4 & 19.3 & 13.3 & 17.4 & 18.1 \\
DocVQA & 50.9 & 65.6 & 70.3 & 48.1 & 63.3 & 71.3 \\
DocVQA-Hard & 30.9 & 45.3 & 46.1 & 30.8 & 41.6 & 48.5 \\
InfographicVQA & 25.8 & 35.5 & 40.3 & 24.4 & 38.8 & 40.6 \\
MMMU & 21.1 & 24.6 & 26.6 & 19.3 & 24.9 & 29.0 \\
TextVQA & 62.9 & 71.1 & 73.4 & 64.5 & 71.5 & 73.2 \\
\bottomrule
\end{tabular}
\caption{\textbf{Token budget} for a 2B model trained on 26B to 1.3T tokens for data-parallel and Decoupled DiLoCo $M=8$. Our method assimilates better than the baseline increased amount of tokens during its training.}
\label{tab:overtrain}
\end{table*}

\section{Infrastructure Details}

\subsection{Deterministic Replay via Event Tapes}\label{sec:tape_generation}

\paragraph{Non-Determinism and Event Logging.}
Decoupled DiLoCo's availability in the event of failures is based on allowing bounded levels of nondeterminism in the training algorithm.
Because the syncer aggregates updates as soon as a minimum quorum $K$ is reached (see \autoref{alg:syncer}), the exact subset of learners used in any given outer optimization step depends entirely on unpredictable wall-clock timing, network jitter, and hardware failures. Additionally, the exact step on which a learner applies a global fragment update to its local model depends on both random failures and the relative speed of the learner and syncer steps at that moment.  To isolate algorithmic behavior from this system noise, the syncer records an \emph{event tape}, $\mathcal{T}$, during training. This tape captures the full causal state of the system by logging the following metadata on each worker at each local step: a vector clock~\citep{mattern1989virtual} which encodes the exact communication pattern between workers as described in Section~\ref{sec:infrastructure}, per-learner token counters, and failure/recovery events.

\paragraph{Deterministic Replay.}
Given a recorded tape $\mathcal{T}$, the syncer and learners can completely bypass the dynamic quorum logic. By reading the tape, the syncer deterministically executes the exact same sequence of synchronization decisions, participant subsets, and token weightings, and each learner applies fragment updates from the syncer and fails or recovers on the same local steps. This mechanism guarantees bitwise-identical training trajectories regardless of the actual hardware conditions during the replay run.

\paragraph{Synthetic Tape Generation.}
Beyond replaying historical runs, we built a discrete-event simulator to generate \emph{synthetic tapes} for controlled ``what-if'' experiments such as chaos engineering for LLMs as described in Section~\ref{sec:simulating_failures}. The generator models the learners and syncer as state machines advancing through virtual time, allowing us to inject configurable disruptions. These include constant speed heterogeneity, transient slowdowns, and realistic large-scale hardware failures modeled either on arbitrarily chosen values as in this paper, or calibrated on production cluster data. Using the deterministic replay execution mode in which each worker faithfully follows the synthetic tapes, we can rigorously evaluate the system's goodput and algorithmic resilience under specific, perfectly reproducible chaotic conditions.

\subsection{Consistent distributed checkpointing}\label{sec:checkpointing}

One of the challenges of moving from a single-controller programming model to a model in which multiple workers execute independently is creation of consistent checkpoints. Although the resilience of Decoupled DiLoCo makes a full system restart less likely, resuming from a checkpoint remains an important fallback strategy in extreme failure scenarios outside the bounds defined by the algorithm. Additionally, in the case of deterministic replay where a specific set of learners must participate in each outer optimization, resuming from a checkpoint remains the main resilience strategy in the case of a failure.

Checkpointing must not block training progress, so each of the workers in the system checkpoints its own state asynchronously to its own directory. One naive strategy would be for each worker to checkpoint independently when its local step meets the criteria $t_w \bmod T_c = 0$. Unfortunately, this strategy provides no guarantee of capturing training progress, because if the causal dependencies between worker states are not captured in the set of checkpoints, resuming from these checkpoints could put the system in an invalid state. Finding a set of uncoordinated checkpoints from all workers that form a consistent global state \citep{mattern1989virtual} can require rolling back all the way to the initial state of the system \citep{elnozahy2002rollbackrecovery}.

\begin{figure}[ht!] 
    \centering
    \resizebox{\columnwidth}{!}{%
        \begin{tikzpicture}[
            >=stealth,
            msg/.style={->, thick, gray!70},
            msgblue/.style={->, thick, blue!60},
            lbl/.style={fill=white, inner sep=2pt, font=\small},
            tick/.style={thick},
            savebox/.style={draw=blue!80, thick, fill=white, rounded corners=2pt, inner sep=3pt, font=\sffamily\small}
        ]
        
        \def\Lzero{0}
        \def\S{4.5}
        \def\Lone{9}
        
        \def\yA{-2}
        \def\yB{-6}
        \def\yC{-10}
        \def\yD{-14}
        \def\bottom{-15.5}
        
        \draw[thick, gray] (\Lzero, 0) -- (\Lzero, \bottom);
        \draw[thick, gray] (\S, 0) -- (\S, \bottom);
        \draw[thick, gray] (\Lone, 0) -- (\Lone, \bottom);
        
        \node[draw, fill=white, minimum width=2.5cm, minimum height=0.8cm, font=\bfseries] at (\Lzero, 0) {Learner 0};
        \node[draw, fill=red!20, minimum width=2.5cm, minimum height=0.8cm, font=\bfseries] at (\S, 0) {Syncer};
        \node[draw, fill=white, minimum width=2.5cm, minimum height=0.8cm, font=\bfseries] at (\Lone, 0) {Learner 1};
        
        \draw[tick] (\Lzero-0.2, \yA) -- (\Lzero+0.2, \yA);
        \draw[tick] (\S-0.2, \yA) -- (\S+0.2, \yA);
        \draw[tick] (\Lone-0.2, \yA) -- (\Lone+0.2, \yA);
        
        \node[lbl, left=0.2cm] at (\Lzero, \yA) {$\langle$L0: 10, L1: 8, S: 8$\rangle$};
        \node[lbl, right=0.2cm] at (\S, \yA) {$\langle$L0: 9, L1: 9, S: 9$\rangle$};
        \node[lbl, right=0.2cm] at (\Lone, \yA) {$\langle$L0: 8, L1: 10, S: 8$\rangle$};
        
        \draw[msg] (\Lzero, \yA) -- (\S, \yB);
        \draw[msg] (\S, \yA) -- (\Lzero, \yB);
        \draw[msg] (\Lone, \yA) -- (\S, \yB);
        \draw[msg] (\S, \yA) -- (\Lone, \yB);

        \draw[tick] (\S-0.2, \yB) -- (\S+0.2, \yB);
        \draw[tick] (\Lone-0.2, \yB) -- (\Lone+0.2, \yB);
        
        \node[lbl, right=0.2cm] at (\S, \yB) {$\langle$L0: 10, L1: 10, \colorbox{yellow!50}{S: 10}$\rangle$};
        \node[lbl, right=0.2cm] at (\Lone, \yB) {$\langle$L0: 9, L1: 11, S: 9$\rangle$};
        
        \node[savebox, above right=0.2cm] at (\S, \yB) {Start Snapshot};

        \draw[msg] (\S, \yB) -- (\Lzero, \yC);
        \draw[msgblue] (\Lone, \yB) -- (\S, \yC);
        \draw[msg] (\S, \yB) -- (\Lone, \yC);

        \draw[tick] (\Lzero-0.2, \yC) -- (\Lzero+0.2, \yC);
        \draw[tick] (\S-0.2, \yC) -- (\S+0.2, \yC);
        \draw[tick] (\Lone-0.2, \yC) -- (\Lone+0.2, \yC);
        
        \node[lbl, left=0.2cm] at (\Lzero, \yC) {$\langle$L0: 11, L1: 10, \colorbox{yellow!50}{S: 10}$\rangle$};
        \node[lbl, right=0.2cm] at (\S, \yC) {$\langle$L0: 10, L1: 11, S: 11$\rangle$};
        \node[lbl, right=0.2cm] at (\Lone, \yC) {$\langle$L0: 9, L1: 12, \colorbox{yellow!50}{S: 10}$\rangle$};
        
        \node[savebox, above left=0.2cm] at (\Lzero, \yC) {Save Checkpoint};
        \node[savebox, above right=0.2cm] at (\Lone, \yC) {Save Checkpoint};

        \draw[msgblue] (\Lzero, \yC) -- (\S, \yD);
        \draw[msg] (\S, \yC) -- (\Lzero, \yD);
        \draw[msgblue] (\Lone, \yC) -- (\S, \yD);
        \draw[msg] (\S, \yC) -- (\Lone, \yD);

        \draw[tick] (\Lzero-0.2, \yD) -- (\Lzero+0.2, \yD);
        \draw[tick] (\S-0.2, \yD) -- (\S+0.2, \yD);
        \draw[tick] (\Lone-0.2, \yD) -- (\Lone+0.2, \yD);
        
        \node[lbl, left=0.2cm] at (\Lzero, \yD) {$\langle$L0: 12, L1: 11, S: 11$\rangle$};
        \node[lbl, right=0.2cm] at (\S, \yD) {$\langle$L0: 11, L1: 12, S: 12$\rangle$};
        \node[lbl, right=0.2cm] at (\Lone, \yD) {$\langle$L0: 10, L1: 13, S: 11$\rangle$};
        \node[savebox, above right=0.2cm] at (\S, \yD) {Save Checkpoint};

        \end{tikzpicture}%
    }
    \caption{Using vector clocks to communicate training progress and coordinate checkpointing. The arrows shown in blue represent messages that will be saved in the syncer's checkpoint and replayed, which may be a different number of messages from each learner when $K < M$.}
    \label{fig:checkpointing}
\end{figure}
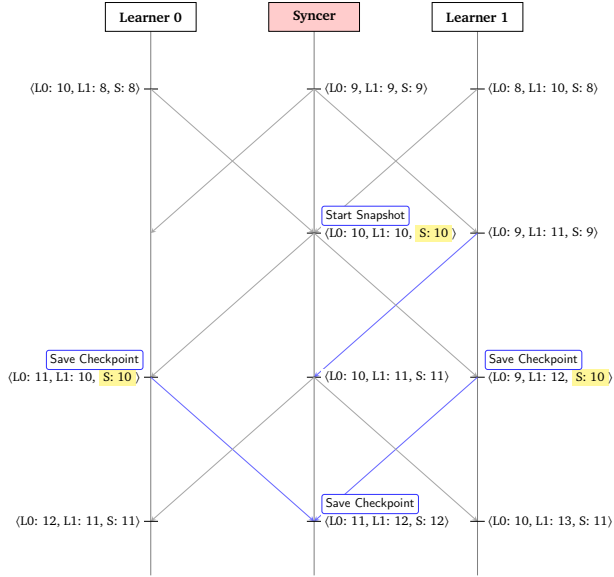

To prevent losing an unbounded amount of progress when a failure occurs, the workers follow a version of the Chandy-Lamport distributed snapshotting algorithm \citep{chandy1985distributed}, as shown in \autoref{fig:checkpointing}. The syncer begins a snapshot when its local step $t_s \bmod T_c = 0$, recording its current model parameters and outer optimizer state. The syncer proceeds with training as usual, sending messages that include its current vector clock to the learner workers. When a learner receives a vector clock from the syncer for which the syncer step meets the condition $t_s \bmod T_c = 0$, this message serves as the snapshot marker \citep{chandy1985distributed}. Receiving this snapshot marker indicates to learner workers that they should checkpoint their own model parameters and inner optimizer state. Since learner workers update their own vector clocks with the latest syncer step, the next message sent from each learner to the syncer serves to return the snapshot marker, communicating to the syncer that the learner has taken its own snapshot. For each learner, there may have been some messages that were not yet received by the syncer at the time that the syncer started its snapshot, which have vector clocks that happen before the marker that the learner recorded its own snapshot. As the syncer receives these messages, it records them in the snapshot. Upon resuming from checkpoint, these messages are replayed by the syncer as if they were resent from the learners, in order to prevent deadlocks which could otherwise occur. The syncer declares its local snapshot complete and writes the snapshot state to the checkpoint directory when it has received a returned marker from every learner. 

To avoid the scenario where a checkpoint cannot complete because some learners fail and never send back the snapshot marker, the syncer omits failed learners from the checkpoint. Upon resumption, instead of loading state from the checkpoint, these failed learners are directed by the syncer to go through the learner recovery process detailed in Section~\ref{sec:recovery}.

\subsection{Distributed Learner Recovery}\label{sec:recovery}

When a learner is slow or temporarily interrupted from training, Decoupled DiLoCo allows it to resume stepping with slightly stale local state and be brought back up to date with the rest of the system through the normal cycle of fragment synchronization. However, when a learner is interrupted for too long, completely fails and restarts, or is newly introduced during compute scavenging as in Section~\ref{sec:exp_scavenging}, resuming contribution to the system requires it to first acquire a recent copy of the model parameters and inner optimizer state. This process is called learner recovery.

The syncer has no special logic to coordinate the recovery of learners - it simply always broadcasts the most recently synchronized fragment to all connected learners. When a recovering learner restarts, it connects to the syncer and waits for the first fragment sync to arrive, stamped with the syncer's latest vector clock $t_s$. It then connects to a healthy peer learner and sends $t_s$ as part of a recovery request. The peer will respond with a copy of its local model parameters and inner optimizer state, but only once it has itself observed the same vector clock $t_s$ coming from the syncer. Until that has happened, it continues training and syncing as normal, leaving the recovery request in a pending state. The transfer of model state from peer learner to recovering learner happens asynchronously but may take significant time, especially if the two learners are geographically distant. During this time the recovering learner buffers any fragment update messages it receives from the syncer, such that when the transfer finishes they can be replayed locally to bring the new learner up to date and able to immediately participate in training. The set of messages received after $t_s$ is guaranteed to be sufficient to fully synchronize the recovering learner, thanks to the peer learner using $t_s$ as a threshold before initiating the transfer.

We apply an upper bound on the number of syncer steps a recovery is allowed to take, equal to the synchronization cycle length $H$. This ensures that the staleness of any fragment and optimizer state in the recovered learner has the same bounds as in any other learner in the system, preventing recovery events from becoming an unpredictable source of instability.

\begin{figure}[ht]
    \centering
    \includegraphics[width=\linewidth]{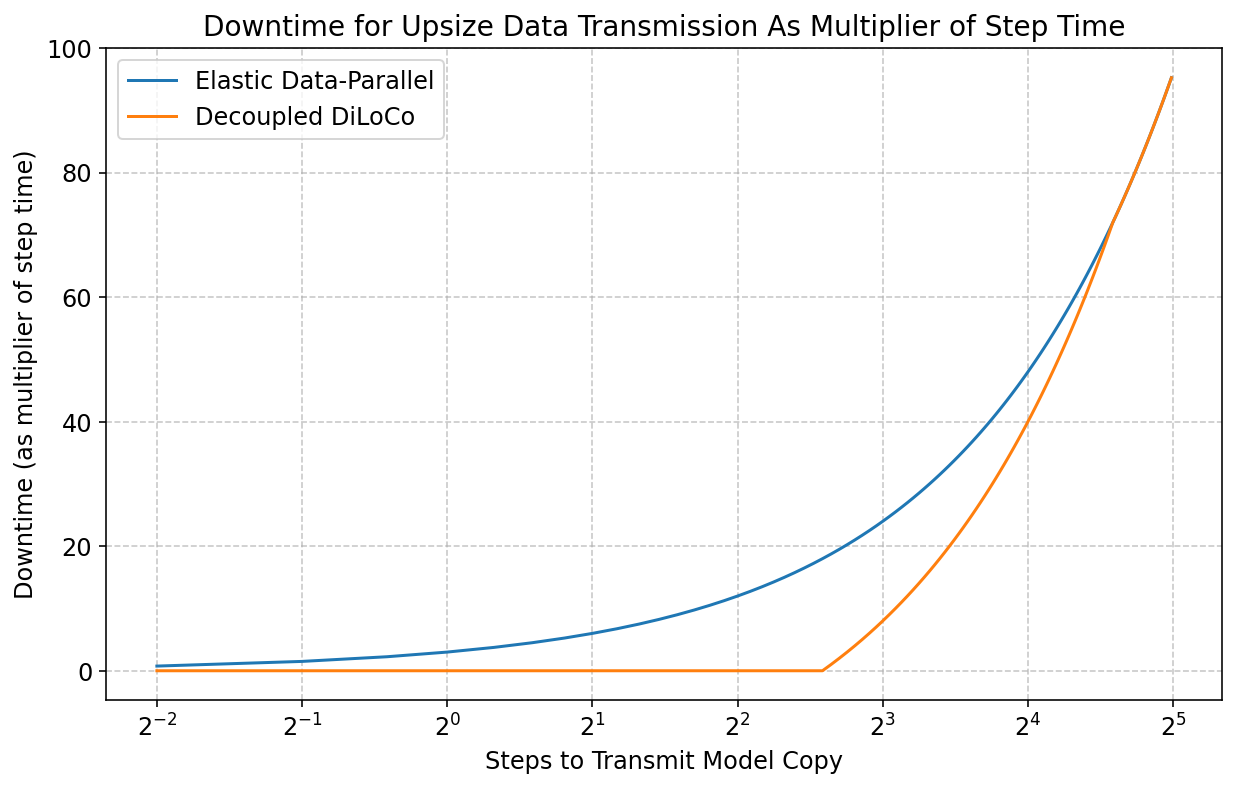}
    \caption{\textbf{Downtime incurred for elastic data-parallel vs Decoupled DiLoCo due to data transmission when upsizing a new replica.} Elastic data-parallel always incurs some overhead for transmitting the most recent model copy, while the overhead for Decoupled DiLoCo is null until the time to transmit a model copy over the available bandwidth increases above $6\times$ the step time (with $H = 24$). When transmission of a single model copy takes $H=24$ steps, Decoupled DiLoCo upsize downtime will catch up to data-parallel.}
    \label{fig:upsize_data_transmission}
\end{figure}

As shown in \autoref{fig:upsize_data_transmission}, the asynchronous design of the learner recovery protocol gives Decoupled DiLoCo an advantage when scavenging with limited bandwidth between compute nodes.  Elastic data-parallel training must block on the transmission of the current model parameters and momentum buffers to the new replica before training, meaning that every upsize event (when a faulty slice has been repaired or replaced), incurs downtime lower-bounded by the time to transmit $3\times$ the model parameters to the new replica. In contrast, Decoupled DiLoCo training is tolerant of the new replica using state up to $H$ steps stale, and transmission of state to the new replica does not block the rest of the system until this limit is reached. As long as there is enough bandwidth available to transmit an additional $\frac{3 \times N}{H}$ model parameters per step, Decoupled DiLoCo will not incur any downtime on an upsize event.

\subsection{Bandwidth profile}\label{sec:bandwidth_profile}

\begin{figure*}[t]
    \centering
    \begin{subfigure}{0.48\textwidth}
        \centering
        \includegraphics[width=\linewidth]{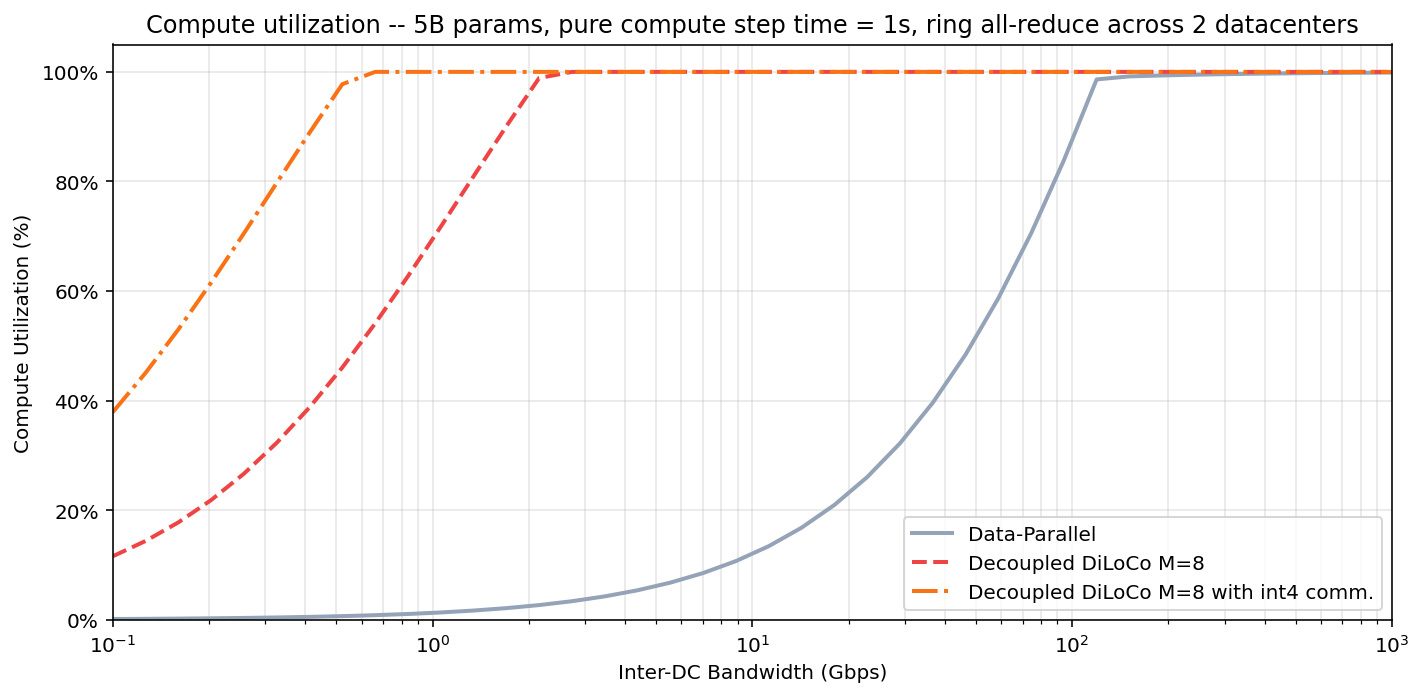}
        \caption{1s compute step time, 2 datacenters}
        \label{fig:bw_1s_2dc}
    \end{subfigure}\hfill
    \begin{subfigure}{0.48\textwidth}
        \centering
        \includegraphics[width=\linewidth]{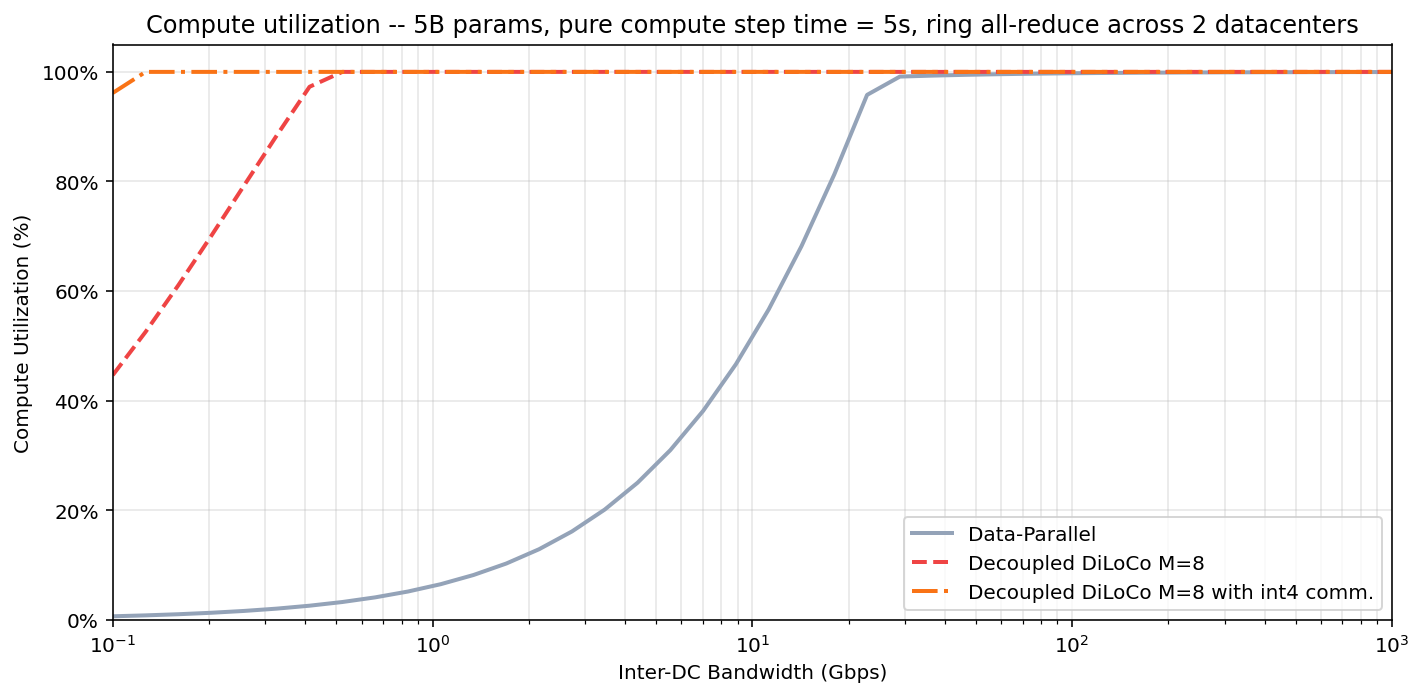}
        \caption{5s compute step time, 2 datacenters}
        \label{fig:bw_5s_2dc}
    \end{subfigure}

    \vspace{1em} 

    \begin{subfigure}{0.48\textwidth}
        \centering
        \includegraphics[width=\linewidth]{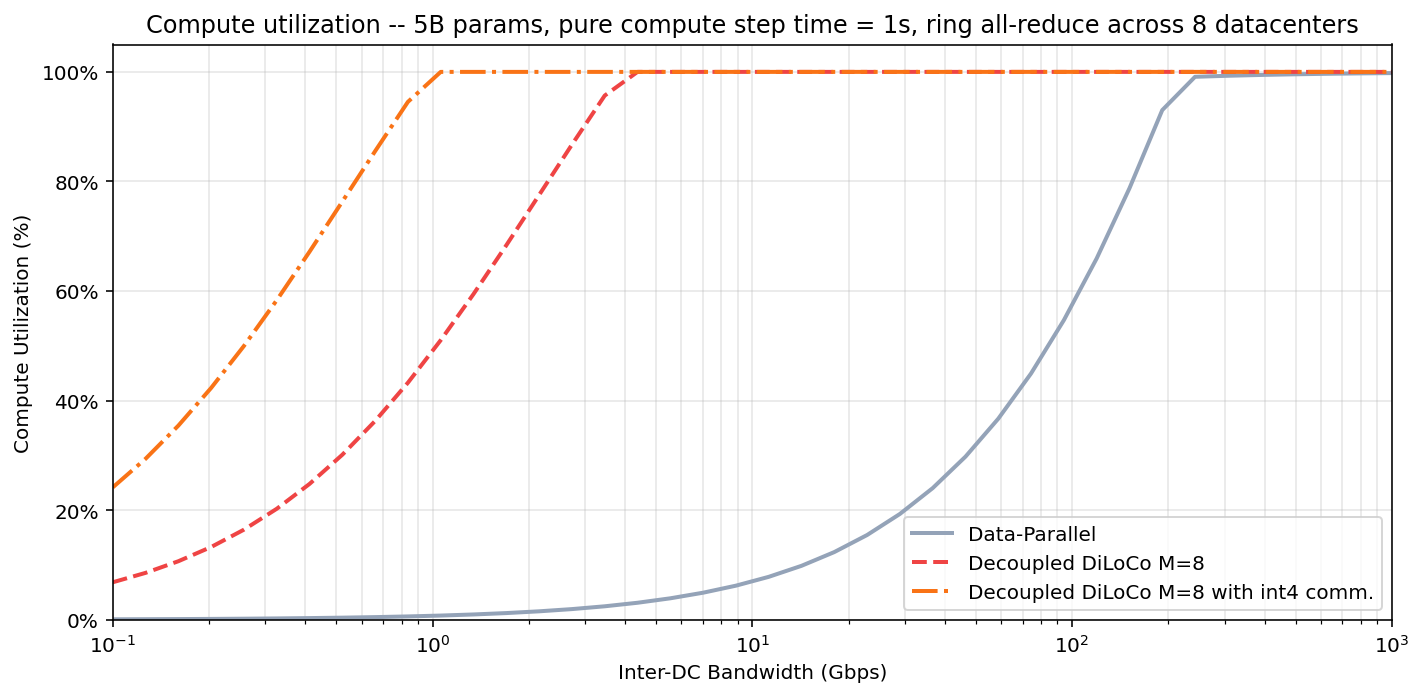}
        \caption{1s compute step time, 8 datacenters}
        \label{fig:bw_1s_8dc}
    \end{subfigure}\hfill
    \begin{subfigure}{0.48\textwidth}
        \centering
        \includegraphics[width=\linewidth]{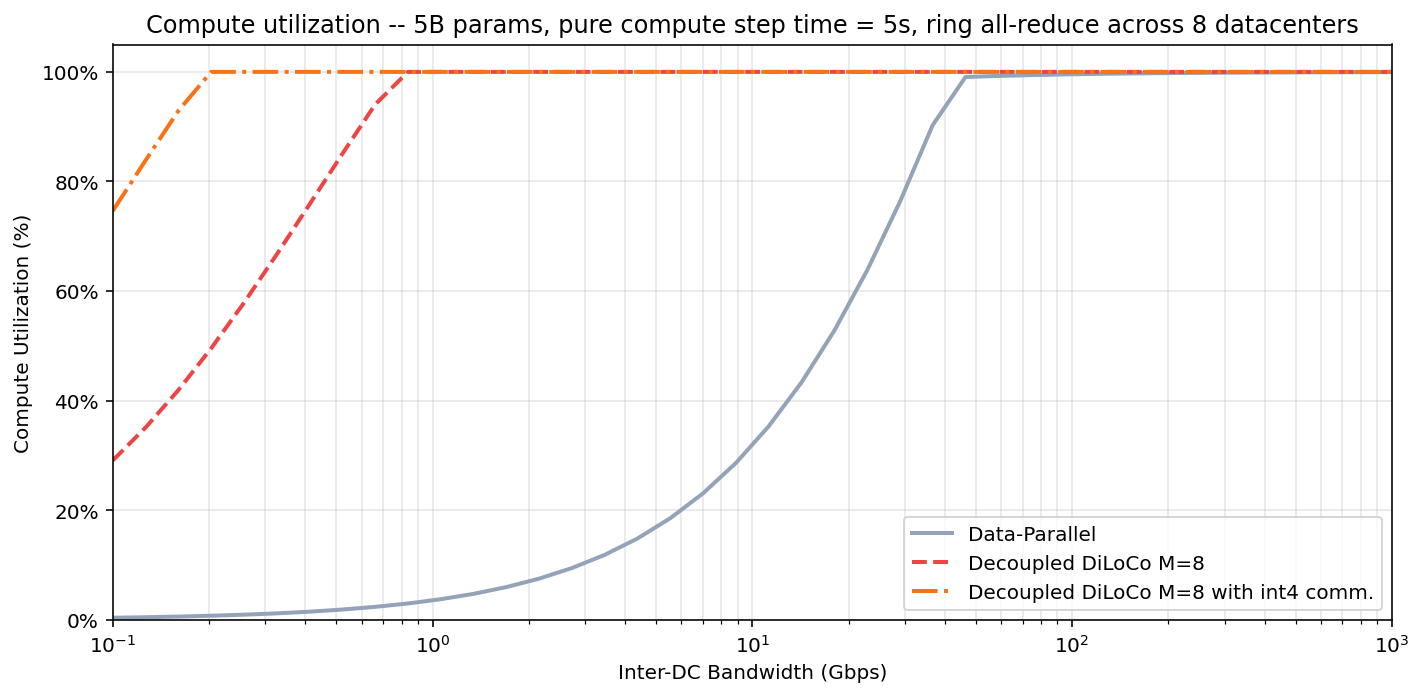}
        \caption{5s compute step time, 8 datacenters}
        \label{fig:bw_5s_8dc}
    \end{subfigure}
    
    \caption{\textbf{Compute utilization} for the 5B parameter model across varying compute step times and datacenters over a range of bandwidth values in Gbits/s.}
    \label{fig:bw_profiles_all}
\end{figure*}

\paragraph{Simulated bandwidth profile} \label{sec:bandwidth_profile_simulated} Following \citet{douillard2025streaming}, we define ``compute utilization'' as the percentage of time doing computation vs doing communication during the all-reduce of the (inner/outer) gradients: $\nicefrac{T_\text{math}}{T_\text{math}+T_\text{comm.}}$. We show in \autoref{tab:compute_utilization_all} and \autoref{fig:bw_profiles_all} the required bandwidth in Gbits/s to reach a certain level of compute utilization, across two step times (1s and 5s) and two cluster sizes (2 datacenters and 8 datacenters). Longer step time $T_\text{math}$ mechanically improves the compute utilization of all methods, and further allows Decoupled DiLoCo to overlap its communication over the computation. More datacenters on which to perform the ring all-reduce will increase the required bandwidth too by a factor of $\nicefrac{2(|\text{DCs}|-1)}{|\text{DCs|}}$ with $\text{DCs|}$ being the number of datacenters on the ring.

In these simulations, Decoupled DiLoCo, with either bf16 (16 bits) or int4 (4 bits) communication size, requires orders of magnitude less bandwidth than its data-parallel counterpart. This massively less need of bandwidth is a critical property for scavenging new resources (see Section~\ref{sec:exp_scavenging}) which can now exploit crumbs of compute in location where bandwidth was not properly allocated beforehand.

\begin{table*}[t]
    \centering
    \begin{subtable}{0.48\textwidth}
        \centering
        \resizebox{\linewidth}{!}{
            \begin{tabular}{lrrrrr}
            \toprule
            \textbf{Method} & \textbf{50\% CU} & \textbf{75\% CU} & \textbf{90\% CU} & \textbf{95\% CU} & \textbf{99\% CU} \\
            \midrule
            Data-Parallel & 49 & 81 & 104 & 113 & 195 \\
            Decoupled DiLoCo M=8 & 0.58 & 1.2 & 1.7 & 1.9 & 2.2 \\
            Decoupled DiLoCo M=8 with int4 comm. & 0.15 & 0.29 & 0.43 & 0.49 & 0.53 \\
            \bottomrule
            \end{tabular}
        }
        \caption{1s compute step time, 2 datacenters}
        \label{tab:5b_1s_2dc}
    \end{subtable}\hfill
    \begin{subtable}{0.48\textwidth}
        \centering
        \resizebox{\linewidth}{!}{
            \begin{tabular}{lrrrrr}
            \toprule
            \textbf{Method} & \textbf{50\% CU} & \textbf{75\% CU} & \textbf{90\% CU} & \textbf{95\% CU} & \textbf{99\% CU} \\
            \midrule
            Data-Parallel & 9.5 & 16 & 21 & 23 & 49 \\
            Decoupled DiLoCo M=8 & 0.12 & 0.23 & 0.34 & 0.39 & 0.43 \\
            Decoupled DiLoCo M=8 with int4 comm. & 0.03 & 0.06 & 0.08 & 0.10 & 0.11 \\
            \bottomrule
            \end{tabular}
        }
        \caption{5s compute step time, 2 datacenters}
        \label{tab:5b_5s_2dc}
    \end{subtable}

    \vspace{1em} 

    \begin{subtable}{0.48\textwidth}
        \centering
        \resizebox{\linewidth}{!}{
            \begin{tabular}{lrrrrr}
            \toprule
            \textbf{Method} & \textbf{50\% CU} & \textbf{75\% CU} & \textbf{90\% CU} & \textbf{95\% CU} & \textbf{99\% CU} \\
            \midrule
            Data-Parallel & 85 & 140 & 183 & 198 & 391 \\
            Decoupled DiLoCo M=8 & 1.0 & 2.0 & 3.0 & 3.3 & 3.8 \\
            Decoupled DiLoCo M=8 with int4 comm. & 0.26 & 0.51 & 0.75 & 0.84 & 0.94 \\
            \bottomrule
            \end{tabular}
        }
        \caption{1s compute step time, 8 datacenters}
        \label{tab:5b_1s_8dc}
    \end{subtable}\hfill
    \begin{subtable}{0.48\textwidth}
        \centering
        \resizebox{\linewidth}{!}{
            \begin{tabular}{lrrrrr}
            \toprule
            \textbf{Method} & \textbf{50\% CU} & \textbf{75\% CU} & \textbf{90\% CU} & \textbf{95\% CU} & \textbf{99\% CU} \\
            \midrule
            Data-Parallel & 17 & 28 & 37 & 40 & 49 \\
            Decoupled DiLoCo M=8 & 0.21 & 0.40 & 0.59 & 0.68 & 0.75 \\
            Decoupled DiLoCo M=8 with int4 comm. & 0.05 & 0.10 & 0.15 & 0.17 & 0.19 \\
            \bottomrule
            \end{tabular}
        }
        \caption{5s compute step time, 8 datacenters}
        \label{tab:5b_5s_8dc}
    \end{subtable}
    
    \caption{Bandwidth requirements in Gbits/s to reach a certain level of compute utilization for the 5B parameter model under varying pure compute step times and datacenter scales.}
    \label{tab:compute_utilization_all}
\end{table*}

\begin{figure*}[ht]
    \centering
    \includegraphics[width=1.0\textwidth]{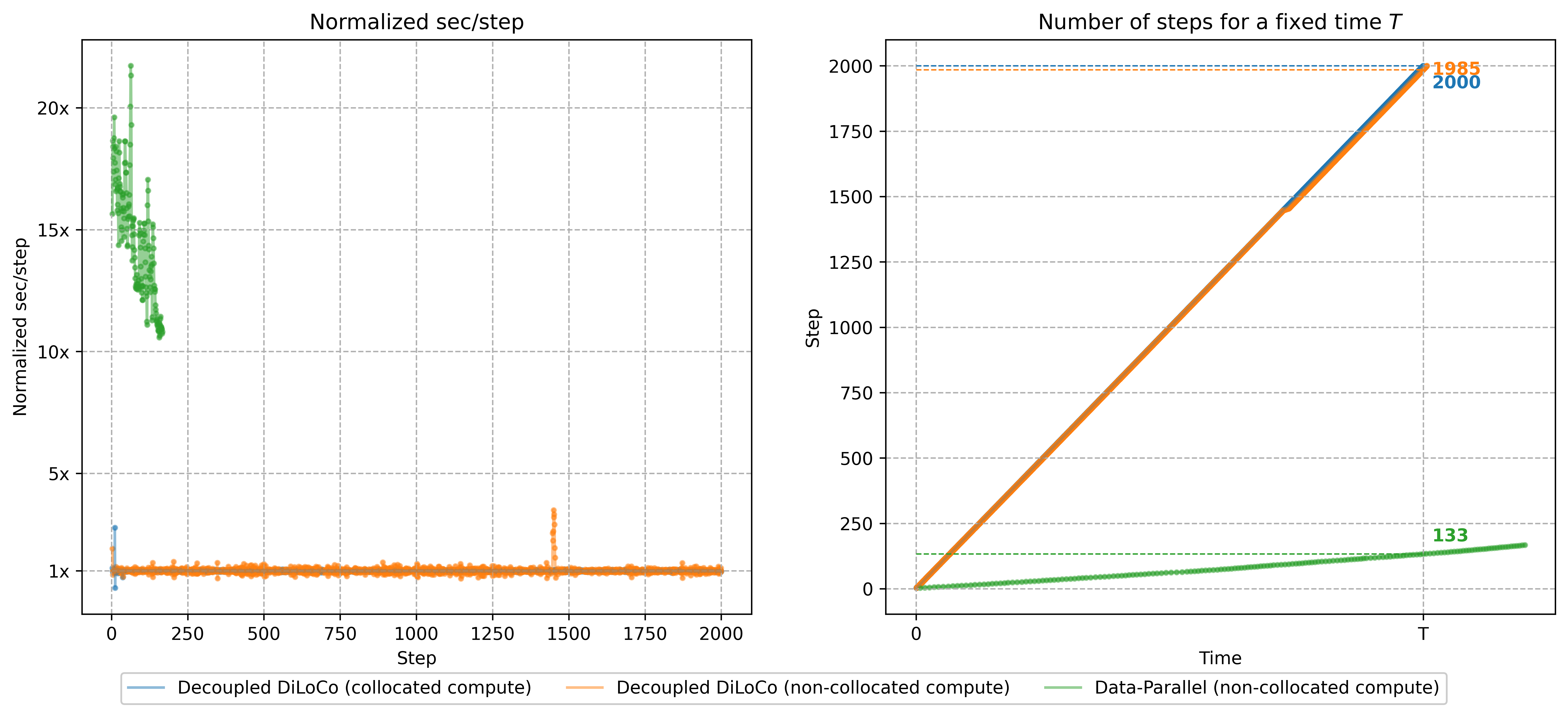}
    \caption{\textbf{Normalized step time and step count to completion} of a non-collocated data-parallel vs with Decoupled DiLoCo $M=8$ both non-collocated and collocated.}
    \label{fig:step_times}
\end{figure*}

\paragraph{Non-collocated Learners} \label{sec:bandwidth_profile_real} We now perform for real a distributed experiment on a custom Chinchilla-like \citep{hoffmann2022trainingcomputeoptimallargelanguage} dense architecture with 12B parameters. We distribute this experiment with $M=8$ learners across the USA, one in the Midwest, two in the South, three out West, and two in the Great Plains. We display in \autoref{fig:step_times} the step time, normalized by the speed of a collocated Decoupled DiLoCo experiment (e.g. all learners in the same datacenter). While Decoupled DiLoCo with non-collocated compute (in orange) is around the same speed as its counterpart with collocated compute (in blue), data-parallel with non-collocated compute (in green) is significantly slower (>10-20x), to the point of being unusable. Note that in this experiment, we didn't make any effort to allocate more bandwidth than the base minimum available and is a somewhat contrived comparison not representative of large-scale effort where proper allocation of bandwidth would be ensured in advance. It is, however, clear evidence that Decoupled DiLoCo can enable more flexible training, for example to scavenge compute across regions.

\begin{figure*}[ht]
    \centering
    \includegraphics[width=1.0\textwidth]{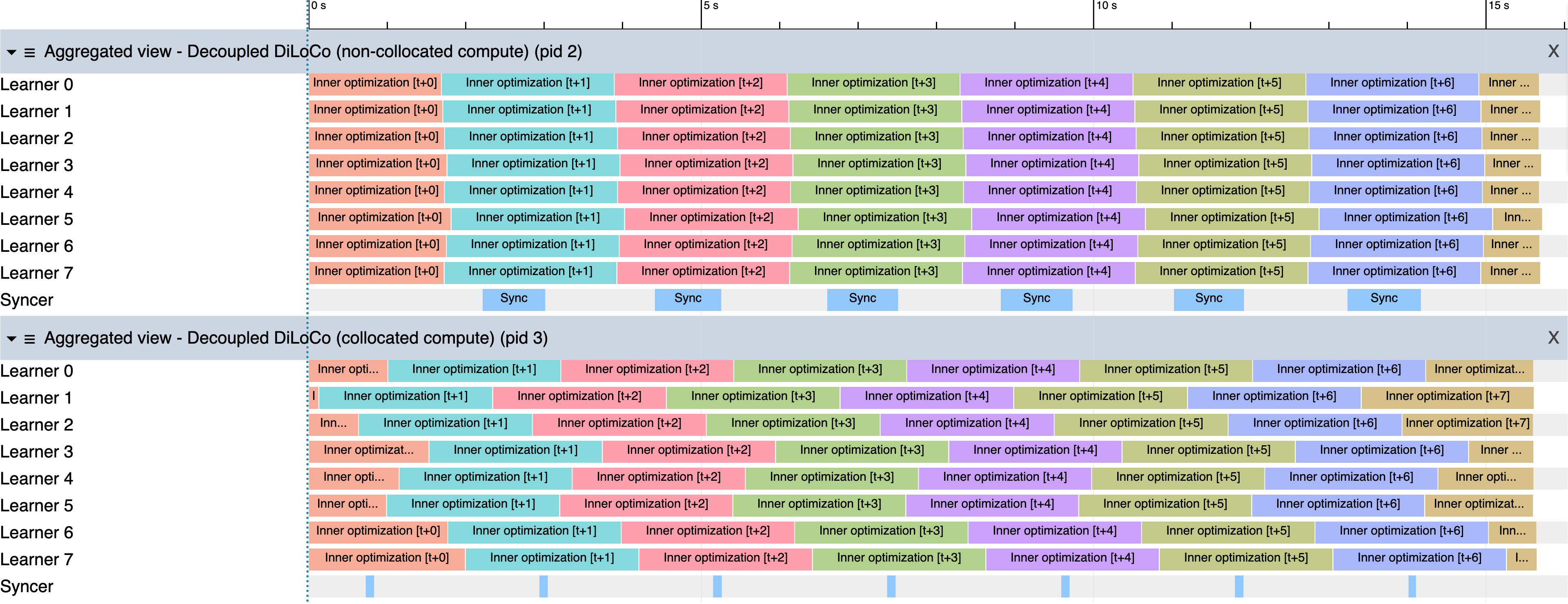}
    \caption{Visualization of the XLA operations done for Decoupled DiLoCo $M=8$ in both non-collocated (top) and collocated (bottom) situations.}
    \label{fig:xprof}
\end{figure*}

We also show in \autoref{fig:xprof} a XProf \citep{openxla_xprof}, representing the real XLA operations done by Decoupled DiLoCo with non-collocated compute (top) vs with collocated-compute (bottom). When compute is collocated, available bandwidth is significantly higher, and thus the syncer time (in shaded blue, at the bottom row) is smaller than with non-collocated compute. However, even in the distributed case, that synchronization time can reliably fit under an entire compute step, and be therefore fully hidden, leading to optimal compute utilization.

\section{Evaluation Benchmarks}
\label{sec:appendix_evals}

To provide a comprehensive assessment of our models, we evaluate performance across a diverse suite of text and vision benchmarks. The datasets used are detailed below.

\subsection{Text Benchmarks}
Our text evaluation suite targets reasoning, commonsense, and reading comprehension capabilities:
\begin{itemize}
    \item \textbf{ARC (Challenge and Easy):} Evaluates question-answering capabilities using grade-school science questions~\citep{clark2018think}.
    \item \textbf{BoolQ:} Tests reading comprehension with yes/no questions derived from search queries~\citep{clark-etal-2019-boolq}.
    \item \textbf{HellaSwag:} Assesses commonsense natural language inference and the ability of models to choose the most logical continuation of a text~\citep{zellers-etal-2019-hellaswag}.
    \item \textbf{PIQA:} Focuses on physical commonsense reasoning, testing a model's understanding of physical interactions~\citep{bisk2020piqa}.
    \item \textbf{SIQA (SocialIQA):} Measures commonsense reasoning specifically regarding social interactions and human behavior~\citep{sap-etal-2019-social-iqa}.
    \item \textbf{WinoGrande:} An adversarial benchmark designed to test pronoun resolution and general commonsense reasoning~\citep{sakaguchi2020winogrande}.
\end{itemize}

\subsection{Vision Benchmarks}
To measure multimodal understanding, we utilize the following vision-based evaluation datasets:
\begin{itemize}
    \item \textbf{ChartQA:} Tests visual and logical reasoning capabilities through question answering based on charts and graphs~\citep{masry-etal-2022-chartqa}.
    \item \textbf{COCO-Captions:} Assesses the model's ability to generate accurate and descriptive captions for everyday images~\citep{chen2015microsoft}.
    \item \textbf{DocVQA:} Evaluates visual question answering specifically on images of document pages \citep{mathew2021docvqa}.
    \item \textbf{InfographicVQA:} A benchmark focused on visual question answering based on complex infographics~\citep{mathew2022infographicvqa}.
    \item \textbf{MMMU:} A massive multi-discipline multimodal evaluation benchmark designed to test expert-level artificial general intelligence~\citep{yue2024mmmu}.
    \item \textbf{TextVQA:} Tests the ability of models to read and reason about text explicitly present within an image~\citep{singh2019towards}.
\end{itemize}

\subsection{Full results}

We display in this section all the results shown in the main paper as figures. Find in \autoref{tab:exp_scaling} the table results from \autoref{fig:scaling} across three scales.

\begin{table*}[!ht]
    \centering
    \begin{adjustbox}{width=0.8\textwidth} 
    \begin{tabular}{l cc cc cc}
        \toprule
        & \multicolumn{2}{c}{Gemma 2B} & \multicolumn{2}{c}{Gemma 5B} & \multicolumn{2}{c}{Gemma 9B} \\
        \cmidrule(lr){2-3} \cmidrule(lr){4-5} \cmidrule(lr){6-7}
        Benchmark & DP & DiLoCo & DP & DiLoCo & DP & DiLoCo \\
        \midrule
        \textcolor{gray}{\small\textit{Text benchmarks (avg)}} & \textcolor{gray}{\small 52.3} & \textcolor{gray}{\small 51.6} & \textcolor{gray}{\small 62.5} & \textcolor{gray}{\small 62.8} & \textcolor{gray}{\small 67.8} & \textcolor{gray}{\small 68.0} \\
        Arc-Challenge   & 36.2 & \textbf{36.3} & \textbf{47.3} & 47.2 & \textbf{56.1} & \textbf{56.1} \\
        Arc-Easy        & \textbf{65.1} & 64.0 & 73.6 & \textbf{73.8} & \textbf{81.6} & 80.4 \\
        BoolQ           & \textbf{58.8} & 55.8 & 65.5 & \textbf{68.8} & 71.3 & \textbf{74.7} \\
        HellaSwag       & \textbf{52.1} & 51.4 & \textbf{65.8} & 64.9 & \textbf{72.1} & 72.0 \\
        Piqa            & 71.8 & \textbf{72.5} & \textbf{76.5} & 76.4 & \textbf{79.3} & 78.9 \\
        Siqa            & 46.0 & \textbf{46.5} & \textbf{48.9} & 46.9 & 49.8 & \textbf{50.4} \\
        Winogrande      & \textbf{55.6} & 54.1 & 59.6 & \textbf{61.3} & \textbf{64.2} & 63.8 \\
        \midrule
        \textcolor{gray}{\small\textit{Vision benchmarks (avg)}} & \textcolor{gray}{\small 44.1} & \textcolor{gray}{\small 43.3} & \textcolor{gray}{\small 49.3} & \textcolor{gray}{\small 47.2} & \textcolor{gray}{\small 56.2} & \textcolor{gray}{\small 55.3} \\
        COCO-Captions   & \textbf{63.8} & 62.9 & \textbf{78.9} & 74.7 & \textbf{87.5} & 86.8 \\
        ChartQA         & 12.2 & \textbf{13.3} & \textbf{19.0} & 17.3 & 21.6 & \textbf{23.0} \\
        DocVQA          & \textbf{50.9} & 48.1 & \textbf{65.6} & 64.0 & \textbf{76.1} & 75.1 \\
        DocVQA-Hard     & \textbf{30.9} & 30.8 & \textbf{45.5} & 44.9 & \textbf{53.2} & 52.3 \\
        InfographicVQA  & \textbf{25.8} & 24.4 & \textbf{37.2} & 36.2 & \textbf{45.9} & 44.8 \\
        MMMU            & \textbf{21.1} & 19.3 & \textbf{26.2} & 24.8 & \textbf{32.9} & 31.3 \\
        TextVQA         & 62.9 & \textbf{64.5} & \textbf{72.8} & 68.7 & \textbf{76.2} & 73.7 \\
        \bottomrule
    \end{tabular}
    \end{adjustbox}
    \caption{Performance comparison of data-parallel (DP) vs. Decoupled DiLoCo ($M=8$) across model scales (2B, 5B, 9B).}
    \label{tab:exp_scaling}
\end{table*}

\subsection{Scavenging full results}
\label{sec:appendix_scavenging_results}

The full downstream evaluation results from the experiments in Section \ref{sec:exp_scavenging} are shown in Table~\ref{tab:scavenging_isoflop_downstream}. Additionally, whilst those experiments detail an \emph{iso-FLOPs} regime where total compute spent per run was fixed, Table~\ref{tab:scavenging_isostep_downstream} shows results from an alternative evaluation setting where the total number of steps is fixed instead, and scavenged compute is used to opportunistically improve downstream task performance given a set time budget. In total, 50\% of the steps in each run under the \emph{iso-step} regime had compute increased by the scavenging factor. We observed that validation loss improved predictably at this scale when increasing the available compute during scavenging windows, and Table~\ref{tab:scavenging_isostep_downstream} shows that these gains translated to real improvements on downstream vision tasks. Text tasks did not on average appear to benefit from the increased compute, but this trend is reflected in the DP baseline, so is likely a property not of DiLoCo but of this model scale, dataset and task set more generally.

\begin{table*}[ht]
\centering

\begin{subtable}{\textwidth}
\centering
\setlength{\tabcolsep}{8pt} 
\resizebox{\textwidth}{!}{
\begin{tabular}{l cccc cccc}
\toprule
\multirow{3}{*}{Metric} & \multicolumn{4}{c}{Data Parallel (DP)} & \multicolumn{4}{c}{DiLoCo} \\
\cmidrule(lr){2-5} \cmidrule(lr){6-9}
& \multicolumn{4}{c}{Scavenging} & \multicolumn{4}{c}{Scavenging} \\
\cmidrule(lr){2-5} \cmidrule(lr){6-9}
& +25\% & +50\% & +100\% & +300\% & +25\% & +50\% & +100\% & +300\% \\
\midrule
\textit{Text Benchmarks (Avg)} & 60.7 & 60.9 & 60.8 & 52.3 & 61.0 & \textbf{61.3} & 61.1 & 60.8 \\
Arc-Chall.      & 43.9 & 45.6 & 44.5 & 31.9 & \textbf{46.3} & 45.4 & 43.8 & 44.0 \\
Arc-Easy        & 70.6 & 70.7 & 71.8 & 57.9 & \textbf{73.2} & 72.2 & 72.5 & 70.9 \\
BoolQ           & 65.6 & 65.8 & 65.7 & 61.9 & 63.5 & 66.6 & \textbf{66.9} & 66.1 \\
HellaSwag       & 61.3 & 61.9 & 62.5 & 46.3 & 62.0 & 62.3 & \textbf{62.7} & 61.9 \\
Piqa            & 75.6 & 75.4 & 75.4 & 69.2 & \textbf{76.1} & 75.6 & 75.6 & 75.1 \\
Siqa            & \textbf{48.3} & 48.0 & 47.5 & 45.0 & 48.1 & 47.6 & 47.9 & 48.0 \\
Winogrande      & 59.4 & 59.0 & 58.2 & 54.1 & 58.0 & 59.2 & 58.5 & \textbf{59.6} \\
\midrule
\textit{Vision Benchmarks (Avg)} & 46.3 & 47.4 & 48.3 & 30.6 & 45.9 & 46.7 & \textbf{48.8} & 47.2 \\
MMMU            & 22.7 & 22.8 & 25.2 & 11.7 & 21.8 & \textbf{25.9} & 24.2 & 25.3 \\
COCO-Cap.       & 74.8 & 75.7 & 75.0 & 52.0 & 71.9 & 73.3 & \textbf{77.6} & 76.7 \\
ChartQA         & 16.5 & 18.2 & 16.6 & 9.8 & 16.2 & \textbf{18.6} & 17.3 & 15.6 \\
DocVQA          & 63.3 & 65.5 & 67.3 & 41.1 & 64.4 & 63.0 & \textbf{67.4} & 63.6 \\
DocVQA-Hard     & 41.1 & 43.8 & \textbf{45.9} & 22.8 & 42.5 & 40.4 & 44.7 & 44.8 \\
InfographicVQA  & 34.2 & 34.6 & 36.9 & 20.3 & 34.6 & 35.5 & \textbf{38.1} & 34.3 \\
TextVQA         & 71.7 & 71.4 & 71.4 & 56.7 & 70.1 & 70.5 & \textbf{72.2} & 70.3 \\
\midrule
\textit{Validation loss delta} & $-0.22\%$ & $-0.47\%$ & $-0.79\%$ & $\mathbf{-1.41\%}$ & $-0.17\%$ & $-0.42\%$ & $-0.78\%$ & $-1.30\%$ \\
\bottomrule
\end{tabular}
}
\caption{\textbf{Iso-step scavenging performance.} All runs take the same number of steps, so higher scavenging factors translate to greater FLOPs and tokens seen.}
\label{tab:scavenging_isostep_downstream}
\end{subtable}

\bigskip 
\vspace{1em}

\begin{subtable}{\textwidth}
\centering
\setlength{\tabcolsep}{8pt}
\resizebox{\textwidth}{!}{
\begin{tabular}{l cccc cccc}
\toprule
\multirow{3}{*}{Metric} & \multicolumn{4}{c}{Data Parallel (DP)} & \multicolumn{4}{c}{DiLoCo} \\
\cmidrule(lr){2-5} \cmidrule(lr){6-9}
& \multicolumn{4}{c}{Scavenging} & \multicolumn{4}{c}{Scavenging} \\
\cmidrule(lr){2-5} \cmidrule(lr){6-9}
& +25\% & +50\% & +100\% & +300\% & +25\% & +50\% & +100\% & +300\% \\
\midrule
\textit{Text (Avg)} & 60.5 & 60.0 & 60.1 & 60.4 & 60.8 & 60.7 & \textbf{60.9} & 60.7 \\
Arc-Chall.      & 42.7 & 44.0 & 43.9 & 43.0 & 44.0 & \textbf{44.8} & 44.1 & 43.1 \\
Arc-Easy        & 70.2 & 70.0 & 72.0 & 71.5 & 70.9 & \textbf{72.8} & 70.5 & 72.0 \\
BoolQ           & 66.8 & 63.4 & 64.6 & 66.1 & 66.1 & 67.2 & 66.4 & \textbf{67.7} \\
HellaSwag       & 60.9 & 60.7 & 61.2 & 60.2 & \textbf{61.9} & 61.0 & 61.2 & 60.9 \\
Piqa            & 75.1 & \textbf{75.8} & 75.7 & 75.2 & 75.1 & 75.7 & 75.1 & 74.4 \\
Siqa            & \textbf{48.6} & 47.5 & 46.9 & 48.1 & 48.0 & 48.2 & 47.5 & 48.2 \\
Winogrande      & 59.1 & 58.6 & 56.6 & 58.6 & 59.6 & 59.1 & \textbf{60.1} & 58.4 \\
\midrule
\textit{Vision (Avg)} & 46.6 & 45.9 & 47.0 & 46.4 & \textbf{47.2} & 45.2 & 46.4 & 46.1 \\
MMMU            & 23.8 & 22.2 & 25.3 & \textbf{25.4} & 25.3 & 25.3 & 22.8 & 20.3 \\
COCO-Cap.       & 76.6 & 73.4 & 71.8 & 71.7 & \textbf{76.7} & 74.4 & 71.6 & 74.6 \\
ChartQA         & 14.8 & 16.0 & 16.8 & \textbf{17.3} & 15.6 & 16.2 & 16.3 & 14.1 \\
DocVQA          & 62.9 & 64.1 & 66.1 & 64.2 & 63.6 & 60.0 & 65.2 & \textbf{66.8} \\
DocVQA-Hard     & 42.4 & 42.6 & 43.8 & 43.7 & \textbf{44.8} & 37.9 & 44.5 & 42.3 \\
InfographicVQA  & \textbf{35.9} & 33.7 & 35.2 & 33.4 & 34.3 & 32.6 & 33.6 & 34.6 \\
TextVQA         & 69.8 & 69.5 & 70.1 & 69.1 & 70.3 & 70.1 & \textbf{70.4} & 70.1 \\
\midrule
\textit{Validation loss delta} & $+0.02\%$ & $\mathbf{-0.04\%}$ & $-0.02\%$ & $+0.07\%$ & $+0.09\%$ & $-0.01\%$ & $+0.08\%$ & $+0.23\%$ \\
\bottomrule
\end{tabular}
}
\caption{\textbf{Iso-FLOPs scavenging performance.} All runs share identical total tokens and theoretical FLOPs.}
\label{tab:scavenging_isoflop_downstream}
\end{subtable}

\caption{\textbf{Scavenging experiment downstream performance} in iso-steps and iso-FLOPs regimes.}
\end{table*}

\end{document}